\documentclass[pdflatex,sn-mathphys-num,iicol]{sn-jnl}

\usepackage{graphicx}%
\usepackage{multirow}%
\usepackage{amsmath,amssymb,amsfonts}%
\usepackage{amsthm}%
\usepackage{mathrsfs}%
\usepackage[title]{appendix}%
\usepackage{xcolor}%
\usepackage{textcomp}%
\usepackage{manyfoot}%
\usepackage{booktabs}%
\usepackage{algorithm}%
\usepackage{algorithmicx}%
\usepackage{algpseudocode}%
\usepackage{listings}%
\usepackage{subcaption}
\usepackage{caption}
\usepackage{xfrac}
\usepackage{nicefrac}
\usepackage{comment}

\theoremstyle{thmstyleone}%
\theoremstyle{thmstyletwo}%

\theoremstyle{thmstylethree}%

\begin{document}

\title[Article Title]{Koopman Eigenfunction-Based Identification and Optimal Nonlinear Control of Turbojet Engine}
\subtitle{\small{Accepted for publication in \textit{Springer Nonlinear Dynamics}.}}


\author*[1]{\fnm{David} \sur{Grasev}}\email{david.grasev@unob.cz}

\affil*[1]{\orgdiv{Department of Aviation Technology}, \orgname{University of Defence}, \city{Brno}, \state{Czech Republic}}




\abstract{Gas turbine engines are complex and highly nonlinear dynamical systems. Deriving their physics-based models can be challenging because it requires performance characteristics that are not always available, often leading to many simplifying assumptions. This paper discusses the limitations of conventional experimental methods used to derive component-level and locally linear parameter-varying models, and addresses these issues by employing identification techniques based on data collected from standard engine operation under closed-loop control. The rotor dynamics are estimated using the sparse identification of nonlinear dynamics. Subsequently, the autonomous part of the dynamics is mapped into an optimally constructed Koopman eigenfunction space. This process involves eigenvalue optimization using metaheuristic algorithms and temporal projection, followed by gradient-based eigenfunction identification. The resulting Koopman model is validated against an in-house reference component-level model. A globally optimal nonlinear feedback controller and a Kalman estimator are then designed within the eigenfunction space and compared to traditional and gain-scheduled proportional-integral controllers, as well as a proposed internal model control approach. The eigenmode structure enables targeting individual modes during optimization, leading to improved performance tuning. Results demonstrate that the Koopman-based controller surpasses other benchmark controllers in both reference tracking and disturbance rejection under sea-level and varying flight conditions, due to its global nature.}

\keywords{Jet engine dynamics, Koopman operator, SINDy, Eigenfunctions, System identification, Nonlinear optimal control}

\maketitle

\section{Introduction}\label{sec1}
Gas turbine engines (GTEs) are highly nonlinear and control nonaffine systems, posing challenges for identification and control system design. The nonlinearities in GTE dynamics mainly result from aero-thermodynamic effects in turbomachinery components, ducts, volumes, combustion processes, and more \cite{bib1}. A major category of GTE modeling and identification methods involves physics-based models. Among these, component-level models (CLMs) are popular, formed by connecting zero-dimensional models of engine components. These are usually represented by their performance characteristics, also known as maps, which are either measured on test stands or computed via computational fluid dynamics (CFD). These models can be used to derive linear parameter-varying (LPV) models for control design or employed directly as on-board estimators. The on-board real-time adaptive CLM was developed in \cite{bib2} and further extended to limit protection \cite{bib3}, model predictive control (MPC) with extended Kalman filter (EKF) \cite{bib4}, and direct thrust control \cite{bib5}. An on-board CLM with an unscented Kalman filter was studied in \cite{bib6}. This combination demonstrated high estimation performance, and was further extended to the MPC framework and health parameter estimation \cite{bib7}. A nonlinear dynamic-inversion controller with a dual EKF was proposed in \cite{bib8}, using a simplified physics-based model with approximate component maps. This model was later converted into a control-affine structure, which is broadly seen as a valid approximation of the actual engine dynamics. A sliding mode controller was used to improve the performance of a model-free adaptive controller for a variable-cycle GTE in \cite{bib9}. A performance-seeking controller based on a dual-loop structure was introduced in \cite{bib10}; the inner loop stabilizes the system, while the outer loop uses a CLM to optimize the GTE’s variable geometry, minimizing fuel consumption and addressing the multi-objective nature of GTE control. Results indicated that the controller reduced fuel consumption by 5 $\%$ while maintaining acceptable performance. On-board CLMs often require substantial computational resources, which may not be available on embedded hardware used on smaller flying platforms.

A plethora of GTE control literature is devoted to linear parameter-varying (LPV) models. To construct these models, linear time-invariant (LTI) subsystems are first identified around multiple steady-state points with subsequent interpolation or fitting of the LTI coefficients w.r.t. to some scheduling parameter, e.g., spool speed. For instance, Wei et al. \cite{bib11} combined a piecewise linear model with nonlinear gains to construct an adaptive Wiener model. The results have shown that the Wiener approach can outperform a general describing function from \cite{bib12} and the piecewise model without the nonlinear gains. The model is often converted to a polytopic structure suitable for the design of robust controllers using quadratic programming and linear matrix inequalities as discussed in, e.g., \cite{bib13,bib14,bib15,bib16,bib17}, where \cite{bib16} is a comprehensive review of GTE control strategies. In \cite{bib17}, the authors also introduce deep learning to engine control. This is the next broad class of nonlinear control approaches - the data-driven control.

The need for component characteristics leads to the first major problem: the construction of some engines does not allow one to reliably measure all characteristics. For instance, backward-flow combustion chambers make it almost impossible to install temperature and other sensors, which are necessary for turbine measurements. The CFD estimation of compressor and turbine maps would require a detailed knowledge of the engine geometry which is either proprietary, or difficult to obtain from measurements. Additionally, it is challenging to estimate the possible dimension changes due to the thermal effects. To avoid these issues, the identification of GTEs rather relies on a black-box approach using dynamical response data that can be easily measured. Among the data-driven methods are, e.g., artificial neural networks (ANNs) \cite{bib17,bib18,bib19} and their variants, like recurrent networks, long short-term memory networks \cite{bib20,bib21}, and auto-regressive nonlinear models with exogenous inputs \cite{bib22,bib23}. 

Other methods focus on the discovery of the underlying governing equations of the system from data. Sparse identification of nonlinear dynamics (SINDy) was developed by Brunton et al. at Washington University and first presented in \cite{bib24}. The method looks for an interpretable governing equation approximated by a linear combination of nonlinear functions of states and inputs. Firstly, a library of candidate functions is designed, and the least squares (LS) problem is subsequently solved to obtain optimal weight coefficients. SINDy focuses on the trade-off between accuracy and interpretability of the model via the L1 regularization. The LS problem is usually solved using the LASSO algorithm \cite{bib25}. SINDy was successfully used for the identification of a small turbojet engine by the authors in \cite{bib26,bib27}, where the accuracy of the model was further enhanced using an EKF, leading to a reliable prediction. Although SINDy is a powerful tool, the resulting model is nonlinear. Despite that, the model can still be converted to a control-affine structure and utilized for MPC, design of EKF as mentioned above, feedback linearization, sliding mode control, and more. 

The Koopman operator theory (KOT) represents a way to transform the nonlinear dynamics into a linear framework in an infinite-dimensional lifted state space \cite{bib28}. In practice, a finite-rank truncation is demanded with a suitable set of transformation functions. This allows for the design of optimal controllers in the lifted state space using quadratic programming approaches known from the regulator optimization as demonstrated, e.g., in \cite{bib29,bib30,bib31}. The lifting is performed using arbitrary nonlinear functions called observables, preserving the nonlinear nature. Although the definition of an optimal set of observables is still a problem-specific and open research question, several data-driven approaches exist to formulate a suitable set of observables without resorting to a trial-and-error approach. 

Regarding the KOT, two approaches can be distinguished, as stated in the following list:
\begin{itemize}
    \item Methods based on dynamic mode decomposition (DMD),
    \item Koopman eigenfunction (spectral) methods.
\end{itemize}

The former includes methods such as extended DMD (EDMD) \cite{bib32}, exact DMD \cite{bib33}, generator EDMD \cite{bib34}, residual DMD \cite{bib35}, kernel DMD \cite{bib36}, Hankel DMD using Krylov sequences \cite{bib28}, and many more. The common goal is to find an optimal approximation of the Koopman operator in the form of a finite-rank matrix via solution of a corresponding LS problem with additional constraints. The observables are constructed either by trial and error, building a large library of candidate observables, e.g., radial basis functions (RBFs) or polynomials, with a subsequent model-order reduction, or using some a priori knowledge about the system. Some methods seek to optimize the observables via machine learning \cite{bib37,bib38,bib39,bib40,bib41}. For instance, the KOT with deep dictionary learning was utilized for identification and fault diagnostics of gas turbines in \cite{bib41}. The observables were represented by an ANN with an autoencoder architecture. A deep transfer learning framework was developed to adaptively capture the degraded states of the turbine. The method was key for the extraction of a reliable nonlinear model. However, utilization of the autoencoder, despite being a powerful tool, leads to a complex model with a nonlinear inverse transformation back to the original state space, making the control design particularly challenging. Our previous work in \cite{bib40} presented a way to optimize the observables using metaheuristic global optimization tools for the identification of GTEs. Results pointed to the suitability of the proposed method for the discovery of low-order Koopman models for GTEs.

The eigenfunction methods focus more on the spectral properties of the Koopman operator. Under the assumption that the aforementioned matrix is diagonalizable, the underlying autonomous dynamics can be decomposed using a linear combination of spectral components represented by eigenvalues, eigenfunctions, and eigenmodes. This was thoroughly discussed in literature, e.g., by Igor Mezić \cite{bib42,bib43,bib44}. One of the rather rigorous spectral methods with a strong mathematical foundation is the generalized Laplace analysis \cite{bib28}. The method is based on an iterative computation using time averages of a weighted signal obtained by subtraction of the previously computed modes from the measured trajectory. Korda et al. \cite{bib45} proposed an optimal construction of Koopman eigenfunctions from data using a dense set of trajectories and mapping only the initial conditions to reveal the spatial structure of eigenfunctions. The method needs to be supplied with guessed eigenvalues, which is the main drawback of this approach. The paper also presents ways to optimize the eigenvalue guesses, but it relies on the analytical gradient. These ideas were also extended to the control synthesis in \cite{bib31}, where several methods were proposed, including the implicit SINDy algorithm or a conversion to an equivalent eigenvalue problem for the discovery of eigenfunctions. The method again assumed a priori knowledge of eigenvalues. The work showed that linear quadratic controllers can be effectively designed in eigenfunction coordinates using simple tools, solving the state-dependent algebraic Riccati equation. 

The main challenges connected to the utilization of these methods for GTE identification and control are listed as follows:
\begin{itemize}
    \item The data required for the identification should capture the autonomous behavior of the system. This is difficult to address for GTEs. A solution would be to accelerate the engine to the maximum operating point, cut off the fuel delivery, and measure the run-down curves. However, control systems of some engines include a cooling logic that keeps the rotor spinning at low RPM in order to actively cool hot sections. Deactivation of this logic could lead to a reduction of the remaining useful life of the specific engine. 
    \item Some necessary performance data cannot be obtained due to the architecture of the engine not allowing for the installation of sensors where needed. Thus, the CLM cannot be developed without additional procedures, such as matching generic component maps to the measured transient response of the GTE. 
    \item Experimental derivation of LPV models without CLM is a time-consuming process, leading only to a locally linearized dynamical model. Furthermore, the onboard adaptive nonlinear models and the corresponding controllers are often computationally expensive. This could complicate their usage, e.g. on small unmanned aerial vehicles, where it might not be possible to install and maintain a processing unit with the required computational power.
\end{itemize}

To address these challenges in this paper, the SINDy algorithm was first utilized to discover the underlying governing equation of the modeled single-spool GTE from data. The data can be measured during normal engine operation under closed-loop control without modifying or disconnecting the existing control system. The testing signal should be designed in such a way that the controlled trajectory meets the engine operating limits. 

Since we do not currently have access to a measured engine response, the identification and analysis were conducted using an in-house nonlinear CLM of a generic single-spool GTE. This model was developed in the MATLAB® environment and validated against the GasTurb commercial engine simulation software \cite{bib46}. A control-affine model was considered a valid approximation of the GTE dynamics. Subsequently, the autonomous part of the SINDy model was utilized for the identification of optimal spectral components, yielding a conversion to the Koopman eigenfunction model (KEM). A genetic algorithm (GA) was employed to find optimal eigenvalues using the temporal projection of autonomous trajectories onto a span of exponential trajectories. A gradient-based algorithm was further developed to find the corresponding eigenfunctions. To enhance the prediction accuracy of the KEM, a Kalman observer was designed. Finally, an optimal Koopman Linear Quadratic Gaussian controller with Integral action (K-LQGI) was designed using global optimization methods and compared against the classical LPV proportional and integral (PI) controller, and its gain-scheduled version. 

The rest of the paper is organized as follows: In Section \ref{sec1}, the introduction to engine modeling and control is provided, and the corresponding problems are formulated. In Section \ref{sec2}, the reference physics-based model of the GTE is presented and briefly described. Section \ref{sec3} briefly introduces the Koopman operator. In Section \ref{sec4} and Section \ref{sec5}, the main contributions of this paper, i.e., the SINDy-KEM-based identification and controller design, are presented. Section \ref{sec6} presents the main results of the application of the proposed approach to a generic GTE. The KEM controller is compared to the classical PI and nonlinear LPV-PI controllers, and another innovation of this paper, the SINDy internal model controller, in both sea-level and varying flight conditions. The final Section \ref{sec7} summarizes the contributions and results.

\section{Component-level model of a turbojet engine}\label{sec2}
To validate the proposed approaches, it was necessary to have a full access to all parameters of the reference nonlinear model, and to have the option to connect various controllers for testing. Thus, an in-house CLM of a generic engine was developed in the MATLAB R2022a environment. The CLM was further utilized for the generation of the training and test datasets. To ensure that the model is a good representative of a real GTE, it was validated against the GasTurb 12 commercial simulation software, which is the benchmark for engine modeling \cite{bib46}.

\subsection{Models of GTE components} \label{sec2.1}

\subsubsection{Atmospheric conditions and losses} \label{sec2.1.1}
Atmospheric conditions were modeled according to the International Standard Atmosphere. The pressure losses were considered in the inlet, combustion chamber (CC), exhaust duct, and exhaust nozzle. The CC losses were modeled as a function of the CC inlet corrected mass flow parameter. 

\subsubsection{Compressor}  \label{sec2.1.2}
Compressor model was represented by the compressor map, where the mass flow, pressure ratio, and efficiency were converted to functions of the corrected spool speed and an auxiliary nonlinear coordinate - $\beta_c$ parameter. The detailed explanation of the modified $\beta_c$ interpolation approach is provided in \cite{bib47}. Thus, the mass flow $W_\text{c}$, pressure ratio $\mathrm{\Pi_c}$ and adiabatic efficiency $\eta_c$ were expressed as 
\begin{equation}\label{eq1}
    [W_{c,corr},\Pi_c,\eta_c] = f_c(\beta_c,N_{corr}),
\end{equation}
where $W_{c,corr}$ and $N_{corr}$ are the corrected compressor air mass flow and the corrected spool speed, respectively. For more details about the GTE corrected quantities, the reader is referred to \cite{bib1}. $N_{corr}$ will be defined later.

The compressor torque $\Gamma_c$ was given as
\begin{equation}\label{eq2}
    \Gamma_c = W_c \hspace{0.5mm} C_{p,a} \left( \Pi_c^{\frac{\gamma_a-1}{\gamma_a}} -1 \right)\frac{1}{\eta_c \hspace{0.5mm} N},
\end{equation}
where $W_c$ is now the physical air mass flow, $C_{p,a}$ is the heat capacity of air, and $\gamma_a$ is the adiabatic exponent of air.

\subsubsection{Combustion chamber}  \label{sec2.1.3}
The combustor model consisted of the map of burning efficiency $\eta_b$ as a function of the combustor loading parameter $\Omega_{cc}$ described, e.g., in \cite{bib46,bib48}, and which accounts for the changes in the engine state and flight conditions. Regarding the pressure losses, these were modeled as a function of both the mass flow parameter and the mean temperature relative to the design point value. This way, the effect of heat addition on the decrease of the total pressure recovery was captured. The power balance of the combustion chamber is given as

\begin{equation}\label{eq3}
    \eta_b W_f H_L = W_{cc} \bar{C}_p(T_{cc},T_{t})(T_{t} - T_{cc}), 
\end{equation}
where $\eta_b$, $W_f$ and $H_L$ are the burning efficiency, fuel mass flow and lower heating value of fuel, respectively, and $\bar{C}_p$, $W_{cc}$, $T_{cc}$ and $T_t$ are the mean heat capacity of gases in CC, CC inlet air mass flow, CC inlet total temperature and turbine inlet total temperature, respectively. 

This equation was utilized to determine the $T_t$ for a known $W_f$. The mean capacity $\bar{C}_p$ is a known function of both inlet and outlet conditions in CC, making the equation implicit. Thus, the fixed-point iteration was employed to solve for $T_t$. Since $\eta_b=f_b(\Omega_{cc})$, and $\Omega_{cc}=f_\Omega(T_{cc}, p_{cc}, W_{cc})$, the efficiency can be rewritten as $\eta_b=f_{b,2}(T_{cc}, p_{cc}, W_{cc})$, making it a known explicit function of the CC inlet conditions.

\subsubsection{Turbine} \label{sec2.1.4}
Similarly to the compressor, the turbine was represented by its map. However, there was no need for a coordinate transformation, and the map could be directly interpolated. The turbine corrected gas mass flow $W_{t,corr}$ and the relative temperature difference $\Delta T/T_{t}$ were functions of the turbine corrected spool speed $N_{t,corr}$ and pressure ratio $\Pi_t$ generally expressed as 
\begin{equation}\label{eq4}
    \left[W_{t,corr}, \frac{\Delta T}{T_t}\right] = f_t(N_{t,corr}, \Pi_t).
\end{equation}

The turbine output torque $\Gamma_t$ was given as
\begin{align}\label{eq5}
    \Gamma_t = W_t \hspace{0.5mm} C_{p,g} \hspace{0.5mm} T_t\left(\frac{\Delta T}{T_t} \right) \frac{1}{N},
\end{align}
where $W_t$ is now the physical turbine mass flow and $C_{p,g}$ is the heat capacity of gas. 

\subsubsection{Nozzle}  \label{sec2.1.5}
The nozzle model included the pressure losses due to both friction and contraction. The critical nozzle pressure ratio $\Pi_{n,crit}$ was also accounted for during the exhaust velocity and thrust calculation. The nozzle velocity $v_n$, mass flow $W_n$, and the thrust $F$ are given as
\begin{align}\label{eq678}
    & v_n  = \begin{cases}
       \sqrt{2 C_{p,g} \hspace{0.5mm} T_n \left( 1 - \Pi_n^\frac{1-\gamma_g}{\gamma_g} \right) \eta_n} &\text{for } \Pi_n < \Pi_{n,crit} \\ 
        \sqrt{\frac{2\gamma_g}{\gamma_g + 1} R T_n}& \text{for } \Pi_n \ge \Pi_{n,crit}
     \end{cases}, \\
    & W_n = \mu_n v_n \frac{p_n}{R T_n} \left[ 1 - \eta_n \left( 1 - \Pi_n^\frac{1-\gamma_g}{\gamma_g} \right) \right]^\frac{1}{\gamma_g - 1} A_n, \\
    & F = W_n v_n - W_0 M_0 \sqrt{\gamma_a R T_0} + p_{n,out} A_n,
\end{align}
where $\eta_n$, $T_n$ and $p_n$ are the adiabatic efficiency, total temperature and total pressure in the nozzle, respectively, $p_0$ and $T_0$ are the atmospheric pressure and temperature, respectively, $\Pi_n = p_n/p_0$ is the nozzle pressure ratio, $\gamma_g$ is the adiabatic exponent of exhaust gases, $R$ is the universal gas constant, $\mu_n$ denotes the nozzle flow coefficient, $A_n$ is the nozzle outlet area, $W_0$ is the engine inlet air mass flow, $M_0$ is the Mach number of flight, and $p_{n,out}$ is the exhaust static pressure measured directly aft the nozzle ($p_{n,out} > p_0$ for supercritical conditions).

\subsubsection{Shaft dynamics and control system} \label{sec2.1.6}
The shaft dynamics were modeled using the following equation of motion: 
\begin{equation}\label{eq9}
    \frac{\text{d}N}{\text{d}t} = \frac{30}{\pi I} \bigg( \Gamma_t\eta_m - \Gamma_c \bigg),
\end{equation}
where the spool speed $N$ is measured in revolutions per minute (RPM), $I$ is the shaft polar moment of inertia, and $\eta_m$ is the mechanical efficiency, which is usually very close to 1. 

Concerning the fuel system, the response of the true fuel flow $W_f$ to the change of a fuel flow command $v$ from the controller was modeled by the following transfer function:
\begin{equation}\label{eq10}
    G_f(s)=\frac{W_f(s)}{v(s)}=\frac{1}{T_f s + 1},
\end{equation}
where $T_f$ is the time constant of the fuel system. 

The control system consisted of the main spool-speed controller (governor) and fuel flow limiters connected to the min-max protection logic. The fuel flow limits were calculated as functions of the spool speed, accounting for the compressor surge and the turbine over-temperature during acceleration, and the compressor choke and combustor blow-out during deceleration. The clamping method was utilized to suppress the integrator wind-up effect.

\subsection{Transient solver} \label{sec2.2}
For the transient calculation, the constant mass flow (CMF) approach coupled with the Newton-Raphson matching algorithm was utilized. For modeling of single-spool turbojet GTEs, the CMF method is advantageous due to its higher accuracy and suitability for long-term transient simulations, as reported in \cite{bib49,bib50}. For more complex engines, e.g., two-spool turbofans, the method becomes computationally expensive and less accurate due to the unmodeled volume dynamics.

In each time step of the transient computation, the change of the air/gas mass flow induced by the changes of the engine state is assumed to instantaneously propagate through the entire engine. Therefore, in any instance, the mass flow of air/gas entering a particular component must be equal to the mass flow leaving the component. For a single-spool turbojet engine, this matching condition has to be satisfied between the compressor and the turbine, and between the turbine and the nozzle. The goal is to find the values of two state variables of the Newton-Raphson method. These variables are guessed and updated until the convergence, mass flow equality, is reached. These state variables are the compressor $\beta_c$ parameter and the turbine pressure ratio $\Pi_t$. The matching errors are given by differences between mass flows obtained from the compressor and the turbine map, and the corresponding mass flows computed based on previous cycle calculations. To update the state variables, the following equation is utilized: 
\begin{equation}\label{eq11}
    \begin{bmatrix}
        \beta_c \\ \Pi_t
    \end{bmatrix}_{k+1} = \begin{bmatrix}
         \beta_c \\ \Pi_t
    \end{bmatrix}_k - \begin{bmatrix}
        \sfrac{\partial e_{m,c}}{\partial \beta_c} & \sfrac{\partial e_{m,c}}{\partial \Pi_t} \\
        \sfrac{\partial e_{m,n}}{\partial \beta_c} & \sfrac{\partial e_{m,n}}{\partial \Pi_t}
    \end{bmatrix}^{-1}\begin{bmatrix}
        e_{m,c} \\ e_{m,n}
    \end{bmatrix}_k,
\end{equation}
where $e_{m,c}$ and $e_{m,n}$ are the compressor-turbine and turbine-nozzle mass flow errors, respectively. The partial derivatives in the Jacobi matrix were approximated using the small perturbation method and finite differences.

\begin{figure*}[!ht]
    \centering
    \includegraphics[width=0.75\linewidth]{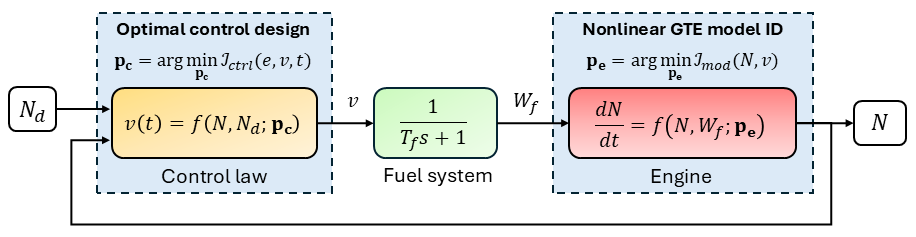}
    \caption{Block diagram of the system}
    \label{block_diagram}
\end{figure*}

The complete system is depicted in Fig. \ref{block_diagram}. It consists of a controller to be designed, the fuel system, which can be accurately represented by the linear model (\ref{eq10}), and the GTE, which is a nonlinear system that needs to be identified from data. The state variables of the GTE and fuel system are the spool speed, $N$, and fuel flow, $W_f$, respectively. The input to the fuel system is the fuel-flow command from the controller. As only the engine rotor dynamics need to be treated as nonlinear, the following identification procedures and transformations will be applied only to the spool speed dynamics. The linear dynamics of the fuel system will be later augmented with the identified GTE model in the Koopman observable subspace.

\section{Koopman operator theory} \label{sec3}
The KOT provides a powerful framework for the representation of nonlinear dynamical systems in a new coordinate system, where the dynamics become linear. The operator itself is theoretically infinite-dimensional. In practice, however, the finite-rank approximation is of interest, and an entire plethora of tools and methods for its discovery were discussed in the literature. One category are the DMD-based methods described in the introduction. In this paper, another approach based on a direct identification of Koopman eigenfunctions was considered. 
Consider the following nonlinear continuous-time dynamical system:
\begin{equation}\label{eq12}
    \mathbf{\dot{x}} = \textbf{f}(\textbf{x},u),
\end{equation}
where $\textbf{x}$ and $u$ are the state and input variables, respectively.

Furthermore, consider a set of nonlinear, Lipschitz continuous, and Lebesgue integrable observable functions: 
\begin{align}
    \mathbf{g}(\textbf{x},u)&=\begin{bmatrix}
    \mathbf{\Psi(x)} & \mathbf{\tilde{u}}(\textbf{x},u)\end{bmatrix}, \nonumber \\[6pt] 
    \mathbf{\tilde{u}}(\textbf{x},0)&=\textbf{0}, \nonumber
\end{align}
 representing the Koopman lifting functions. Thus, it is assumed that the unforced dynamics can be described using only the observables $\mathbf{\Psi}(\textbf{x})$. The associated Koopman operator $\mathcal{K}^t$ is defined by the following composition equation:
\begin{equation}\label{eq13}
    \mathcal{K}^t\mathbf{g} = \mathbf{g} \circ \mathbf{S}^t,
\end{equation}
where $\mathbf{S}^t$ is the flow map capturing the evolution of system states after time $t$, which parametrizes the flow.  

Note that there exists an entire family of Koopman operators parameterized by $t$ (sampling period). To present eigenfunctions, the Koopman generator form is more convenient. When $t \rightarrow 0$, the infinitesimal change of observables is given by a Lie derivative as
\begin{equation}\label{eq14}
    \lim_{t \rightarrow0} \frac{\mathcal{K}^t\mathbf{g - g}}{t} = \mathcal{L}\textbf{g} = \dot{\textbf{g}} = \nabla\textbf{g} \cdot \textbf{f},
\end{equation}
where $\mathcal{L}$ is the Koopman infinitesimal generator. 

Infinitely many Koopman operators $\mathcal{K}^t$ can be sampled from a single generator given a time step $t$. For the unforced dynamics, special observables called eigenfunctions are particularly of interest. The Koopman eigenfunction $\varphi(\textbf{x})$ and the corresponding eigenvalue $\lambda$ are defined as \cite{bib28}
\begin{align}\label{eq15}
    &\mathcal{K}^t\varphi=\varphi\circ \mathbf{S}^t = e^{\lambda t}\varphi, \nonumber \\
    &\mathcal{L}\varphi = \lambda \varphi,  
\end{align}
which thus directly defines the eigenfunction dynamics as $\dot{\varphi} = \lambda\varphi$ with solution $\varphi(t)=\varphi(0) e^{\lambda t}$.

This allows for a spectral decomposition of an observable of interest to a linear combination of $(\lambda,\varphi,c)$ tuples given as
\begin{equation}\label{eq16}
    g(\textbf{x}) = \sum_{i=1}^n c_i \varphi_i(0) e^{\lambda_i t},
\end{equation}
where $c$ are the mode amplitudes, or simply eigenmodes, representing a projection of the observable onto a span of eigenfunctions.

In practice, the infinite-dimensional operator $\mathcal{K}$ is approximated by a matrix $\textbf{K}$, and associated eigen-tuples obtained from data-driven methods such as EDMD. For a system with inputs, this matrix can be further split as $\mathbf{K = \begin{bmatrix} \mathbf{A} & \textbf{B} \end{bmatrix}}$, where $\textbf{A}$ and $\textbf{B}$ are the system and input matrices, respectively. The eigenfunctions and modes can be obtained by diagonalization of $\textbf{A}$ as it represents the unforced (autonomous) dynamics. EDMD is one of the common methods for the data-driven identification of Koopman systems \cite{bib32}. The continuous-time Koopman linear system is thus given as
\begin{equation}\label{eq17}
    \mathbf{\dot{\Psi} = A\Psi +B\tilde{u}}.
\end{equation}

In EDMD, a library of nonlinear observable functions is first constructed for the state dynamics lifting. Subsequently, the finite-rank matrix $\textbf{K}$ is obtained by LS minimization and represents an optimal mapping from one time step to the next across the entire dataset. Finally, diagonalization of the system matrix $\textbf{A}$ is carried out to obtain eigenvalues and eigenfunctions as
\begin{align}\label{eq18}
    \mathbf{\dot{\Psi}} &= \mathbf{A\Psi = V\Lambda V^{-1} \Psi}, \nonumber \\
    \mathbf{V^{-1} \dot{\Psi}} &= \mathbf{V^{-1} V \Lambda V^{-1} \Psi}, \nonumber\\
    \mathbf{\dot{\Phi}} &= \mathbf{\Lambda \Phi}.
\end{align}

However, as these eigenfunctions are constrained to linear combinations of the same basis, $\mathbf{\Psi}$, the limitations are already given by the selection of $\mathbf{\Psi(x)}$. These observables should be designed in such a way that the projection $\mathbf{\Phi=V^{-1} \Psi}$ is well-defined for the most prominent eigenfunctions, which is not always guaranteed. Furthermore, EDMD might lead to solutions that are ill-conditioned and possibly contain lightly unstable eigenvalues, although the model itself provides a very accurate fit \cite{bib31}. The selection of optimal EDMD observables is often a trial-and-error process, which can be supported by approximation and learning methods as discussed in Section \ref{sec1}. 

In the spectral approach, these prominent eigenfunctions are identified directly from data with no need for the construction of a global set of observables, since the eigenfunctions themselves are the optimal intrinsic coordinates directly connected to the rich underlying dynamics. Moreover, they span an invariant subspace due to their algebraic structure \cite{bib44,bib45}. $\mathbf{\Phi(x)}$ can be expressed as a linear combination of arbitrary basis functions that can adapt to each eigenfunction individually. Thus, there are no constraints given by the projection of eigenfunctions on a common fixed set of observables. 

Taking the Lie derivative of eigenfunctions leads to the Koopman partial differential equation (KPDE) given as
\begin{equation}\label{eq19}
    \mathbf{\dot{\Phi}=\nabla\Phi \cdot \dot{x} = \nabla\Phi \cdot f(x},0) \mathbf{=\Lambda \Phi}.
\end{equation}

The KPDE agrees with the generator formulation (\ref{eq14}, \ref{eq15}) and uniquely defines eigenfunctions for a given set of eigenvalues. This is particularly useful for the identification of eigenfunctions from data \cite{bib31}.

\section{Proposed approach - SINDy-KEM}\label{sec4}
Eigenfunctions can be reliably identified for the autonomous dynamics of the system. However, in the case of GTEs, it is rather challenging to measure the autonomous nature of the dynamics as mentioned in Section \ref{sec1}. 

The idea of direct identification of eigenfunctions from data using regression and KPDE originates in the discussed works of Korda et al. \cite{bib45} and Kaiser et al. \cite{bib31}. In this paper, it is further combined with the SINDy algorithm utilized for the estimation of the autonomous dynamics. This enables the discovery of the underlying dynamics from data gathered during normal engine operation. Thus, the extensive experimental procedures, such as linearization around multiple steady-state points for deriving an LPV model, are bypassed. Furthermore, there is no need for measurements of the component characteristics and the engine geometry, both of which are not always accessible, and simplifying assumptions that might compromise the accuracy when chosen inappropriately.

\begin{figure}[!ht] 
    \centering
    \includegraphics[width=1\linewidth]{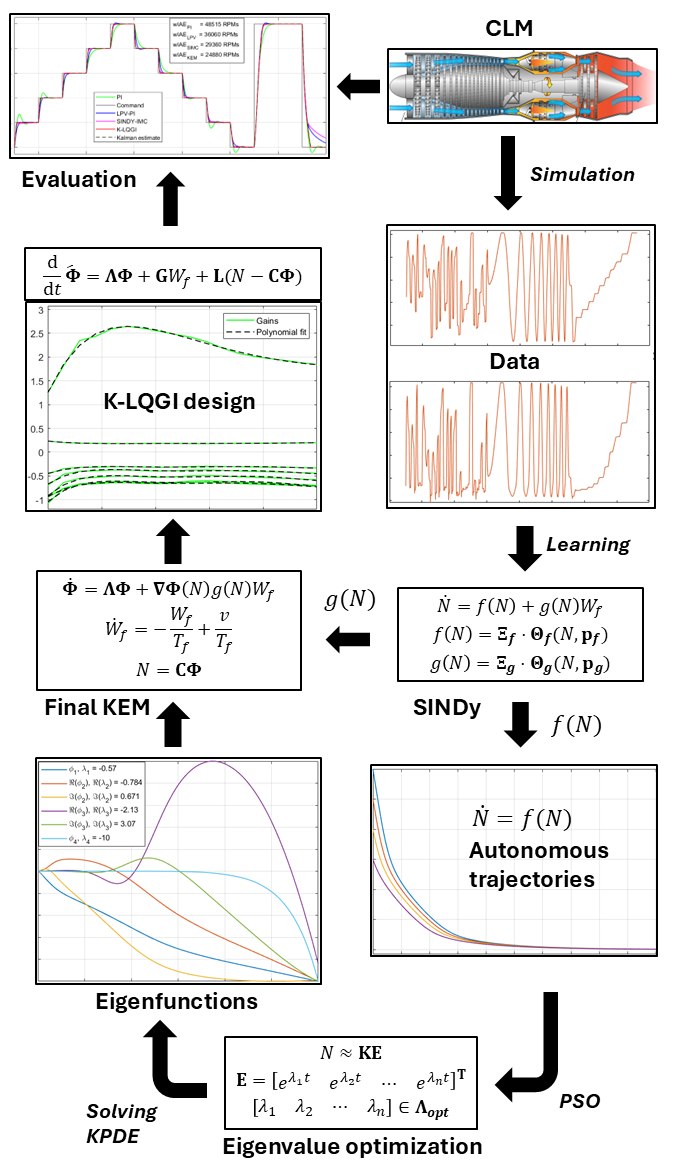}
    \caption{A workflow diagram of the entire identification and control design process}
    \label{fig1}
\end{figure}

The workflow involves identifying the engine governing equation, extracting the estimated autonomous dynamics, discovering the optimal set of eigenvalues, corresponding eigenfunctions, and eigenmodes, and transforming the nonlinear governing equation into the KEM. Subsequently, the control system is designed and optimized. The summary is provided in Fig. \ref{fig1}.

\subsection{Identification of the control-affine governing equation}  \label{sec4.1}
The SINDy model is the most important part of the entire process since it sets the maximum accuracy of the KEM. Thus, it is crucial to properly capture the nonlinear dynamics. The dynamics of single-spool GTEs can be generally described as
\begin{equation}\label{eq20}
    \dot{N} = f(N,W_f,p_{1t},T_{1t}),
\end{equation}
where $p_{1t}$ and $T_{1t}$ are the inlet total pressure and temperature, respectively, capturing varying flight conditions and external effects. 

For a more compact representation, still accounting for varying flight conditions, the corrected quantities are utilized as follows \cite{bib1} (Chapter 8):
\begin{align}
    N_{corr} &= N_{phys}\sqrt{\frac{288}{T_{1t}}}, \label{eq21}\\ 
    W_{f,corr} &= W_{f,phys}\sqrt{\frac{288}{T_{1t}}}\frac{101325}{p_{1t}}, \label{eq22}
\end{align}
where the subscript $(\cdot)_{phys}$ denotes physical quantities.

Let us hereafter use the shorter notation $N$ and $W_f$ for normalized corrected quantities given as
\begin{align}
    N &= \frac{N_{corr} - O_N}{S_N}, \label{eq_norm_N} \\
    W_f &= \frac{W_{f,corr} - O_W}{S_W}, \label{eq_norm_W}
\end{align}
where $O_N$, $O_W$, $S_N$, and $S_W$ are the normalization offsets and scales for spool speed and fuel flow, respectively.

The simplified control-affine model of the GTE is thus given as
\begin{equation}\label{eq23}
    \dot{N}=f(N) + g(N)W_f,
\end{equation}
where $f$ and $g$ are the autonomous and input dynamics, respectively. 

The SINDy cost function is given as
\begin{align}\label{eq24}
    J_{S} = &\left\|\dot{N} - \mathbf{\Xi_f}\mathbf{\Theta_f}(N,\mathbf{p_f}) + \mathbf{\Xi_g}\mathbf{\Theta_g}(N,\mathbf{p_g})W_f \right\|_F^2 \nonumber \\ 
    & + \alpha\| \mathbf{\Xi} \|_1,
\end{align}
where $\mathbf{\Xi}=\begin{bmatrix} \mathbf{\Xi_f} & \mathbf{\Xi_g} \end{bmatrix}$ is the vector of weights, $\mathbf{\Theta_f}$ and $\mathbf{\Theta_g}$ are the autonomous and input candidate functions, respectively, $\mathbf{p_f}$ and $\mathbf{p_g}$, are vectors of nonlinear parameters of candidate functions for $f(N)$ and $g(N)$, respectively, with $\mathbf{p}=\begin{bmatrix} \mathbf{p_f} & \mathbf{p_g} \end{bmatrix}$, and $\alpha$ is the L1 regularization weight factor. 

Since also the nonlinear parameters of basis functions are optimized, this problem is solved using gradient descent (GD) methods with adaptive 
momentum (ADAM) optimizer \cite{bib51}. Additionally, to promote sparsity of the solution, the L1 regularization is combined with sequential thresholding (ST) \cite{bib24}. Whenever any weight in $\mathbf{\Xi}$ falls below a specified threshold, it is excluded from the equation. After each ST round, the GD algorithm is initialized again to fine-tune the remaining active terms in the model. This procedure repeats until a suitable trade-off between the complexity, given by the number of active terms, and accuracy is reached.

The main equations of GD with ADAM are given below:
\begin{equation}\label{eq25}
    \begin{bmatrix}
        \mathbf{\Xi} \\ \textbf{p}
    \end{bmatrix}_{k+1}=
    \begin{bmatrix}
        \mathbf{\Xi} \\ \textbf{p}
    \end{bmatrix}_k -\left( \frac{\alpha \mathbf{v}}{\sqrt{\mathbf{s}+\epsilon}}\right)_{k},
\end{equation}
where 
\begin{equation}    \label{eq26}
\textbf{v}_{k+1}=\beta_1\textbf{v}_{k}+(1-\beta_1)\nabla J_{S,k},
\end{equation}
\begin{equation}\label{eq27}
    \textbf{s}_{k+1}=\beta_2 \textbf{s}_k+(1-\beta_2) {\nabla J_{S,k}}^2,
\end{equation}
\begin{equation}\label{eq28}
    \alpha_k=\eta_p \left(\frac{1-{\beta_2}^k}{1-{\beta_1}^k}\right),
\end{equation}
\begin{equation}\label{eq29}
    \nabla J_{S,k} = \begin{bmatrix}
         \frac{\partial J_{S}}{\partial \mathbf{\Xi}} &  \frac{\partial J_{S}}{\partial \textbf{p}} 
    \end{bmatrix}_k^\intercal, 
\end{equation}
and where $\epsilon$ is an arbitrarily small number, $\beta_1$ and $\beta_2$ are tunable constants usually set as 0.9 and 0.999, respectively, and $\eta_p$ is the baseline learning rate value. 

An additional cost term is included to force the autonomous function to satisfy $f(0)=0$. The term and its gradient are given as
\begin{equation}\label{eq30}
    J_0 = K_0 |f(0)|,
\end{equation}
\begin{equation}\label{eq31}
    \nabla J_0 = K_0 \frac{|f|}{f} \begin{bmatrix}
        \frac{\partial f}{\partial \mathbf{\Xi}} & \frac{\partial f}{\partial \textbf{p}}
    \end{bmatrix}^\intercal \bigg|_{t=0} ,
\end{equation}
where $f = \mathbf{\Xi_f}\mathbf{\Theta_f}(N,\mathbf{p_f})$ and $K_0$ is the weight factor, which should be high enough to enforce the condition, but not too high to deform $f(N)$ in the vicinity of 0, as the learning algorithm might be too focused on $f(0)$. 

The ST is performed according to the following conditions:
\begin{align}
    \xi = 
    \begin{cases}\label{eq32}
        0, \ \text{for } \ |\xi| \le \xi_{min} \\
        \xi, \ \text{for } \ |\xi| > \xi_{min}
    \end{cases},
\end{align}
where $\xi_{min}$ is the threshold, and also according to the properties of the given basis functions. For instance, if logistic functions (LFs) were utilized, a shift of the center too far from the active region would make the function either 0 or 1, which would effectively exclude it or merge it with the other constant functions in the set.

Next step is the extraction of the autonomous dynamics $f(N)$ by setting $W_f=0 \hspace{1mm}kg/s$, and generating a set of autonomous trajectories. The function $f(N)$ is integrated starting from multiple initial conditions until the zero state is reached. This provides an estimate of the true autonomous dynamics of the GTE. It should be noted that the zero normalized state, $N = 0$, corresponds to the minimum sustainable physical spool speed, usually in the ground idle (GI) regime. The GI physical spool speed and fuel flow are subtracted from the corrected data, $N_{corr}$ and $W_{f,corr}$, and the result is scaled for the normalization according to (\ref{eq_norm_N}) and (\ref{eq_norm_W}). In this paper, the normalization was performed using $O_N = 5000$ RPM, $S_N = 10000$, $O_W = 0.04$ kg/s, and $S_W = 0.8$.

The estimated trajectories can be pictured as a measured response of the GTE to an instantaneous step change of the fuel flow from a value corresponding to the initial spool speed to the GI value.

\subsection{Optimization of the eigenvalues via temporal projection} \label{sec4.2}
For the optimal construction from \cite{bib45}, a priori initial guesses of the eigenvalues need to be supplied. The article mentions a possible optimization of eigenvalues, but it assumes a gradient-based approach. However, we want to apply constraints and enforce conditions that must be satisfied by eigenvalues. The penalization terms may not have analytical gradients, or their evaluation might be expensive. 

The goal is to decompose the estimated autonomous dynamics $f(N)$ into a linear combination of Koopman spectral components given as
\begin{equation}\label{eq33}
    f(N) = \sum_{i=1}^n c_i\lambda_i\varphi_i(N) = \sum_{i=1}^n c_i\lambda_i\varphi_i(N_0)e^{\lambda_i t},
\end{equation}
alternatively, to decompose the trajectories generated by $f(N)$, as
\begin{equation}\label{eq34}
    N = \sum_{i=1}^n c_i\varphi_i(N) = \sum_{i=1}^n c_i\varphi_{i}(N_0)e^{\lambda_i t}.
\end{equation}

Equation (\ref{eq33}) represents the spatiotemporal decomposition of the dynamics. Note that due to the properties of the linear combination, the eigenfunctions can be scaled so that the initial conditions are $\varphi_i(N_0)=1$. Thus, the temporal decomposition can be rewritten as
\begin{equation}\label{eq35}
    N = \sum_{i=1}^n k_i(N_0)e^{\lambda_i t} = \textbf{K}\textbf{E},
\end{equation}
where $k_i=c_i\varphi_i, k_i \in \textbf{K}$ are scaled mode amplitudes and $\textbf{E}$ is a matrix of exponential trajectories.  

The KPDE offers a powerful tool for the discovery of the Koopman intrinsic coordinates - eigenfunctions, and thus, no trial-and-error procedure connected to DMD-based methods is needed. The problem to be addressed is finding an optimal set of eigenvalues providing a good basis for the projection of the autonomous dynamics. This set is composed of the most prominent eigenvalues with negative real parts close to the imaginary axis, as these are the longest persisting in the dynamics. Heavily damped eigenvalues, which will decay almost immediately, are not a suitable choice. A possible option is to identify a large set of eigenfunctions and subsequently utilize regularized LS to find a sparse vector of eigenmodes. For this, the algebraic structure of eigenvalues and eigenfunctions can be effectively exploited. However, it can be a time-consuming process that can often lead to many spurious eigen-tuples \cite{bib31}. 

In this paper, a meta-heuristic global optimization method called particle swarm optimization (PSO) was employed to solve the eigenvalue optimization problem since it would be difficult to address the following constraints with gradient-based approaches. This algorithm is a popular choice for global optimization and is supported by the MATLAB Global Optimization Toolbox. For a detailed description, the reader is referred to \cite{bib52,bib53}.  

The single-spool GTE is a 1st-order system. Thus, the problem can be decoupled into a temporal projection of autonomous trajectories onto trajectories generated by a given set of eigenvalues, such that the projection error is minimized, and subsequent discovery of spatial shapes of eigenfunctions. The former should be reached by optimization of the given set of eigenvalues. Trajectories starting from $k$ different initial conditions are augmented to a single matrix $\mathbf{N_{traj}}=\begin{bmatrix} N_1 & N_2 & \cdots & N_k \end{bmatrix}^\intercal$. The set of Koopman modes should be optimal across the entire operating region. The complete constrained problem is formulated as

\begin{equation}\label{eq36}
    \mathbf{\Lambda} = \arg \min_\mathbf{\Lambda} \| \mathbf{N_{traj}}-\textbf{K} \textbf{E}(\mathbf{\Lambda}) \|_F^2,
\end{equation}
s.t.
\begin{align}\label{eq37}
    & \textbf{E}(\mathbf{\Lambda}) = \begin{bmatrix}
        e^{\mathbf{\lambda}_1 t} & e^{\mathbf{\lambda}_2 t} & \cdots & e^{\mathbf{\lambda}_n t}
    \end{bmatrix}^\intercal, \nonumber \\ 
    & \textbf{K}^\intercal=\left(\textbf{E}\textbf{E}^\intercal+\alpha_K\textbf{I}\right)^{-1}\mathbf{E N_{traj}}^\intercal
    \hspace{2mm} \bigg |_{\mathbf{\Lambda}=\text{const.}}, \nonumber  \\
    & \text{Re}(\mathbf{\Lambda})_{\text{min}} \le  \text{Re}(\mathbf{\Lambda})  \le \text{Re}(\mathbf{\Lambda})_{\text{max}} < \textbf{0}, \nonumber \\
    & \text{Im}(\mathbf{\Lambda})_{\text{min}} \le  \text{Im}(\mathbf{\Lambda})  \le \text{Im}(\mathbf{\Lambda})_{\text{max}}, \nonumber \\
    & \frac{\text{Im}(\mathbf{\Lambda})}{\text{Re}(\mathbf{\Lambda})} \le \left(\frac{\text{Im}(\mathbf{\Lambda})}{\text{Re}(\mathbf{\Lambda})}\right)_{\text{max}},
\end{align}
where $\textbf{K}$ is the matrix of Koopman modes associated with the given set of eigenvalues, $\mathbf{\Lambda}$, $\alpha_K$ is a regularization weight factor, and $\text{Re}(\mathbf{\Lambda})_{\text{max}}$ ensures a sufficient stability margin.

The objective function addresses the above-mentioned constraints through penalization. It also includes an innovative additional feature that allows for switching between distinct and repeated real eigenvalues, for which the corresponding generalized eigenfunctions obey the following equations:
\begin{equation}\label{eq38}
    \frac{\text{d}}{\text{d}t}
    \begin{bmatrix} \varphi_R \\ \varphi_D \end{bmatrix} = 
    \begin{bmatrix} \lambda & 1 \\ 0 & \lambda \end{bmatrix}  
    \begin{bmatrix} \varphi_R \\ \varphi_D \end{bmatrix},
\end{equation}
where $\varphi_R$ and $\varphi_D$ denote eigenfunctions corresponding to the distinct and repeated eigenvalues, respectively.

The solution is given as
\begin{equation}\label{eq39}
    \begin{bmatrix}  \varphi_{R} \\ \varphi_{D} \end{bmatrix} = 
    \begin{bmatrix}  e^{\lambda_1 t} & te^{\lambda_1 t} \\ 0 & e^{\lambda_1 t} \end{bmatrix} 
    \begin{bmatrix} \varphi_{R,0} \\ \varphi_{D,0} \end{bmatrix}.
\end{equation}

The $te^{\lambda_1 t}$ term is called the secular term and corresponds to a transient energy growth. In the context of fitting the autonomous dynamics, secular terms can help adapt the model better to the shape of trajectories. 

Generation of purely exponential trajectories (distinct real) or secular terms (repeated eigenvalues) is based on a condition of a minimum distance between two consecutive eigenvalues. Eigenvalues are first sorted in descending order and distances between each subsequent pair are computed as $\mathbf{\Lambda}(2:\text{end})-\mathbf{\Lambda}(1:\text{end}-1)$. If the distance between a given pair of eigenvalues falls below a specified threshold, the old pair is replaced by a new pair of repeated eigenvalues corresponding to the mean of the old pair. 

The exponential trajectories represent a basis, onto which the autonomous trajectory is projected in the time domain. Whenever there are repeated eigenvalues, then not only the purely exponential trajectory $e^{\lambda t}$, but also the secular trajectory $e^{\lambda t}+te^{\lambda t}$ is added to the basis. This way, the optimization algorithm summarized in Algorithm 1 can switch between the cases of distinct and repeated eigenvalues.

Regarding the complex eigenvalues, even though the single-spool engine is a strictly dissipative system (RPM monotonously decreases to 0), the slightly oscillatory modes can help to better explain the underlying dynamics, as the model can mathematically better conform to the shape of autonomous trajectories. Note that the complex eigenvalue is written as $\lambda = \alpha \pm i\beta$. Thus, we use the notation $\text{Re}(\lambda)=\alpha$ and $\text{Im}(\lambda)=\beta$.

To avoid excessively oscillatory modes, the imaginary-to-real-part ratio $\beta/\alpha$ is checked, and the unsatisfactory modes are penalized according to some function $f(\beta/\alpha)$. The higher the imaginary part, the higher the natural frequency of the mode. Thus, if the ratio is too high, the real part, corresponding to damping, does not sufficiently decay the amplitude of oscillations. This is summarized in Algorithm \ref{algo2}. In this paper, the function is
\begin{align}
    f(\beta/\alpha) = (\beta/\alpha)_{max},
\end{align}
exceeding which leads to exclusion of the mode.

Furthermore, the principle of switching to repeated complex eigenvalues can also be adopted, with the distance now being computed in the complex plane. For this purpose, an upper-triangular distance matrix \textbf{D} can be computed, and whenever there is a pair of too close eigenvalues, the secular terms are generated as
\begin{equation}\label{eq40}
   \begin{bmatrix}
        \Re(\varphi_{R}) \\  \Im(\varphi_{R}) \\  \Re(\varphi_{D}) \\  \Im(\varphi_{D})
    \end{bmatrix}
      = e^{\bar{\alpha} t}
      \begin{bmatrix}
        C({\bar{\beta}}) & -S({\bar{\beta}}) & tC({\bar{\beta}}) & -tS({\bar{\beta}})\\
        S({\bar{\beta}}) & C({\bar{\beta}}) & tS({\bar{\beta}}) & tC({\bar{\beta}}) \\
        0 & 0 &C({\bar{\beta}}) & -S({\bar{\beta}}) \\
        0 & 0 &  S({\bar{\beta}}) & C({\bar{\beta}})
    \end{bmatrix},
\end{equation}
where $\bar{\alpha}=(\alpha_i+\alpha_j)/2$ and $\bar{\beta}=(\beta_i+\beta_j)/2$, $S({\bar{\beta}})$ and $C({\bar{\beta}})$ denote $\sin \bar{\beta t}$ and $\cos \bar{\beta t}$, respectively. The initial conditions are all equal to 1, and the distance matrix is given as
\begin{equation}\label{eq41}
    \textbf{D} = \begin{bmatrix}
        0 & d_{1,2} & \cdots & d_{1,n} \\
        0 & \ddots & & \vdots \\
        \vdots &  & \ddots & d_{n-1,n} \\
        0 & \cdots & \cdots & 0
    \end{bmatrix},
\end{equation}
where
\begin{equation}\label{eq42}
    d_{i,j} = \sqrt{\left(\text{Re}(\lambda_i) - \text{Re}(\lambda_j)\right)^2 + \left(\text{Im}(\lambda_i) - \text{Im}(\lambda_j)\right)^2}.
\end{equation}

However, the existence of repeated complex modes is rather rare. Furthermore, the real solutions are also often present even when the complex eigenvalues are considered. To further increase the adaptivity of the algorithm, whenever the imaginary part is smaller than a specified value, the pair of complex conjugate eigenvalues is merged to a single real eigenvalue by setting $\alpha=-\infty$ for one of these eigenvalues. This effectively reduces the order of the system. Alternatively, the pair could be replaced by a pair of repeated real eigenvalues, but that was not considered in this paper.

\begin{algorithm}[h]
    \caption{Objective function - real eigenvalues}\label{algo1}
\begin{algorithmic}
\State \textbf{Input} $\mathbf{\Lambda}$, Auton. traj. data
\State \textbf{Output} $J(\mathbf{\Lambda})$

\vspace{1mm}
\State \textbf{sort}($\mathbf{\Lambda}$)     \Comment{Sorts eigenvalues}
\State $\textbf{dist} = \mathbf{\Lambda}(2:end)-\mathbf{\Lambda}(1:end-1)$

\vspace{2mm}
\For{i = 1 : n-1}
\If{$\textbf{dist}(\lambda_i,\lambda_{i+1})$ $<$  tol.}
    \State $\lambda_i=\lambda_{i+1}=(\lambda_i+\lambda_{i+1})/2$    \Comment{Replaces with the mean $\lambda$}
    \State $\textbf{E}_i=e^{\lambda_i t}$
    \State $\textbf{E}_{i+1}=(1+t)e^{\lambda_{i+1} t}$ \Comment{Secular terms}
\Else
\EndIf
\EndFor

\vspace{3mm}

\State $\textbf{K}^\intercal=\left(\textbf{E}\textbf{E}^\intercal+\alpha_K\textbf{I}\right)^{-1}\textbf{E}\textbf{N}_{\textbf{traj}}^\intercal$ \Comment{Computes projection matrix}
\State $J= \left\|\textbf{N}_{\textbf{traj}}-\textbf{K}\textbf{E} \right\|_F^2$

\end{algorithmic}
\end{algorithm}

\begin{algorithm}[h]
    \caption{Objective function - distinct complex eigenvalues}\label{algo2}
\begin{algorithmic}
\State \textbf{Input} $\mathbf{\Lambda}$, Auton. traj. data
\State \textbf{Output} $J(\mathbf{\Lambda})$

\vspace{1mm}
\State $\alpha=$ $\Re(\mathbf{\Lambda})$
\State $\beta=$ $\Im(\mathbf{\Lambda})$
\State $\textbf{E}=\begin{bmatrix}
    e^{(\alpha \pm i\beta)_1t} &  e^{(\alpha \pm i\beta)_2t} & \cdots &  e^{(\alpha \pm i\beta)_nt}
\end{bmatrix}^\intercal$ 

\vspace{1mm}
\State $\textbf{K}^\intercal=\left(\mathbf{EE^\intercal} +\alpha_K\textbf{I}\right)^{-1}\textbf{E}\textbf{N}_{\textbf{traj}}^\intercal$ \Comment{Computes projection matrix}
\State $P_{oscil}=f(\beta/\alpha)$ \Comment{Penalizes excessively oscillatory modes}

\vspace{1mm}
\State $J=\left\|\textbf{N}_{\textbf{traj}}-\textbf{K}\textbf{E} \right\|_F^2 + P_{oscil}$

\end{algorithmic}
\end{algorithm}

\subsection{Identification of analytical eigenfunctions from KPDE} \label{sec4.3}
 The temporal projection optimization produces optimal eigenvalues and mode amplitudes. Furthermore, by plotting the exponential time trajectories as functions of spool speed, it also reveals the shape of eigenfunctions. However, direct fitting of these shapes was met with problems regarding the accuracy of the final model, especially due to the gradient of eigenfunctions being sensitive to shape discrepancies of the fitted eigenfunctions. Therefore, the KPDE cost function can be minimized instead, and the gradient-based algorithm is utilized for this purpose. The revealed shapes can help with the selection of basis functions for the subsequent regression. 
 
 For a single-spool turbojet GTE, being a 1st-order system, the KPDE (\ref{eq19}) for a single eigenfunction reduces to the following ODE:
\begin{equation}\label{eq43}
    \frac{\partial \varphi}{\partial N} f(N)=\lambda\varphi.
\end{equation}

This equation can be solved analytically as
\begin{equation}\label{eq44}
    \ln{\varphi} = \lambda \int \frac{1}{f(N)} \text{d}N.
\end{equation}

In practice, however, $f(N)$ can be arbitrarily complex, especially for highly nonlinear systems such as the GTE. Thus, since the analytical computation can get extremely challenging, it becomes more advantageous to solve (\ref{eq43}) numerically via the method of candidate solutions closely related to the SINDy algorithm. 

\begin{figure}[!ht] 
    \centering
    \includegraphics[width=1\linewidth]{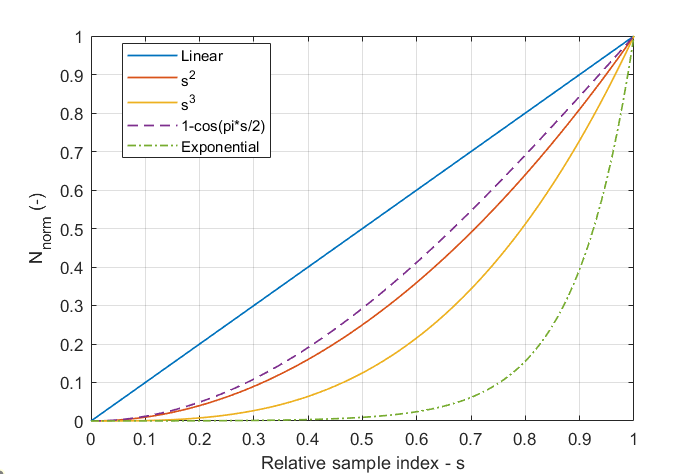}
    \caption{Examples of nonlinear sampling approaches}
    \label{fig2}
\end{figure}

Instead of the utilization of generated autonomous trajectories, the function of autonomous dynamics $f(N)$ is sampled using a nonlinear sampling pattern that ensures an increasing density of sample points towards 0. Examples of nonlinear sampling approaches are provided in Fig. \ref{fig2}, where the horizontal and vertical axes represent the relative position of the sample and the corresponding $N$ value, respectively. 

Similarly to the $f(N)$ function, the eigenfunctions of a dissipative system should also satisfy $\varphi(0)=0$ for all modes. This is because even the longest persisting mode (largest eigenvalue) should vanish after a finite time and thus the $\dot{N} \rightarrow 0$. The nonlinear sampling also indirectly makes the algorithm put more effort into satisfying this condition. 
 
 Additionally, as the optimization is two-sided in the sense that both the gradient and the eigenfunction (both terms in the error) are optimized simultaneously, it is crucial to prevent the learning from heading directly towards the trivial solution $\mathbf{\Xi=0}$. This is avoided by a penalty term that is a function of the difference between the value of the eigenfunction at $N_0$ in the current iteration and the value at $N_0$ in the first iteration. This way, the cost term acts as an artificial spring, which forces the solution to stay away from the origin of the $\mathbf{\Xi}$ parameter space. The larger the distance, the larger the counterforce. Thus, the complete cost function for a single $\varphi=\mathbf{\Xi\Theta}(N,\textbf{p})$ and distinct real eigenvalue $\lambda$ is given as
\begin{align}\label{eq45}
    J_{e,r,d} = &\left\| \mathbf{\Xi} \frac{\partial \mathbf{\Theta}}{\partial N} \dot{N} - \lambda\mathbf{\Xi\Theta} \right\|_2^2+\alpha_1\|\mathbf{\Xi}\|_1 \nonumber\\
                & + \alpha_2 \|\varphi(N_0)_{1}-\varphi(N_0)\|_2^2,
\end{align}
where $\alpha_1$ and $\alpha_2$ are the weights of the L1 sparsity-promoting regularization and the trivial solution penalty, respectively, and $\varphi(N_0)_1$ and $\varphi(N_0)$ represent the initial value of the eigenfunction at the first iteration and the current iteration, respectively.

Concerning the repeated eigenvalues, the first cost term corresponding to the generalized eigenfunction is given by
\begin{equation}\label{eq46}
    J_{e,r,r} = \left\| \mathbf{\Xi}\frac{\partial \mathbf{\Theta}}{\partial N}\dot{N} - \lambda\mathbf{\Xi\Theta} - \varphi_{\lambda,d} \right\|_2^2,
\end{equation}
where $\varphi_{\lambda,d}$ is the eigenfunction corresponding to the distinct eigenvalue.

For the complex eigenfunctions, the KPDE is given as
\begin{equation}\label{eq47}
    \frac{\partial}{\partial N} (\varphi_R \pm i\varphi_I)\dot{N} = (\alpha \pm i\beta)(\varphi_R \pm i\varphi_I).
\end{equation}

Splitting it into the real and the imaginary part, we obtain
\begin{align}
   \text{Re}: \frac{\partial \varphi_R}{\partial N} \dot{N} = \alpha\varphi_R-\beta\varphi_I, \label{eq48} \\
    \text{Im}: \frac{\partial \varphi_I}{\partial N} \dot{N} = \beta\varphi_R+\alpha\varphi_I. \label{eq49}
\end{align}

The first term of the cost function is thus given as
\begin{align} \label{eq50}
   J_{e,c} = &\left\|\frac{\partial \varphi_R}{\partial N} \dot{N} - \alpha\varphi_R+\beta\varphi_I\right\|_F^2 \nonumber \\
             & + \left\| \frac{\partial \varphi_I}{\partial N} \dot{N} - \beta\varphi_R-\alpha\varphi_I\right\|_F^2,
\end{align}
where $\alpha$ and $\beta$ are the real and the imaginary part of the eigenvalue, respectively, and $\varphi_R=\mathbf{\Xi_R\Theta_R}(N,\mathbf{p_R})$ and $\varphi_I=\mathbf{\Xi_I\Theta_I}(N,\mathbf{p_I})$.

\subsection{Autonomous trajectory projection and the complete KEM} \label{sec4.4}
Once the eigenfunctions are identified, the mode amplitudes are obtained by solving the following regularized LS problem in the spatial domain:
\begin{equation} \label{eq51}
    \textbf{C}=\arg\min_{\textbf{C}} \left\| N-\textbf{C}\mathbf{\Phi}(N) \right\|_2^2 + \alpha_C \| \textbf{C} \|_1 ,
\end{equation}
or, equivalently, in the temporal domain:
\begin{equation} \label{eq52}
    \textbf{C}=\arg \min_\textbf{C} \left\| N(t)-\textbf{C}\mathbf{\Phi}(N_0)e^{\mathbf{\Lambda} t} \right\|_2^2 + \alpha_C \| \textbf{C} \|_1 ,
\end{equation}
where $\alpha_C$ is the regularization parameter.

Regularization helps in exclusion of the inactive modes and preserves only the prominent ones, leading to further dimensionality reduction. The final model is given as
\begin{align}  \label{eq53}
    \dot{\mathbf{\Phi}}&=\nabla \mathbf{\Phi}\dot{N} \nonumber \\
    &=\nabla \mathbf{\Phi}\left(f(N) + g(N)W_f\right) \nonumber \\
    &=\mathbf{\Lambda}\mathbf{\Phi}+\nabla\mathbf{\Phi} g(N) W_f,
\end{align} 
where the RHS can be further approximated using a suitable basis, e.g., polynomials or RBF, or converted to a look-up-table format. 

Unlike the locally linearized interpolated LPVs, this system is valid across the entire operating region of the GTE. Furthermore, the only nonlinear component is the input dynamics, which can be sampled by evaluation at arbitrary points in the state space. The sampled subsystems are LTI and coherent to the underlying dynamics. This will be leveraged during the control design.

An example of an affine LPV decomposition of the input dynamics $\nabla\mathbf{\Phi}(N) g(N)$ is presented in Appendix \ref{secA1}. The discrete-time matrices can be obtained as $\mathbf{L}=\text{exp}({\mathbf{\Lambda}t})$ and $\mathbf{G_d = \Lambda^{-1}(L-I)\nabla\Phi}g$.

\section{Controller design} \label{sec5}

\subsection{Proportional-integral controller} \label{sec5.1}
A control system with a PI controller and fuel-flow limiters was designed using the nonlinear CLM. According to the PART 31 regulation, the control system should be able to safely increase the thrust from 15 $\%$ to 95 $\%$ of the nominal value in under 5 seconds \cite{bib54}. For this purpose, the PI controller was manually tuned to achieve the desired performance. 

\subsection{LPV-PI controller} \label{sec5.2}
To simplify the control design in the frequency domain, the SINDy governing equation was linearized analytically using the following equations:
\begin{equation} \label{eq54}
    a(N,W_f) = \frac{\partial f(N)}{\partial N}+\frac{\partial g(N)}{\partial N}W_f,
\end{equation}
\begin{equation} \label{eq55}
    b(N) = \frac{\partial f(N)}{\partial W_f}+\frac{\partial g(N)W_f}{\partial W_f}=g(N),
\end{equation}
and the transfer function in the frequency domain was
\begin{equation}  \label{eq56}
    G_e(s) = \frac{K_e(N)}{T_e(N)s+1},
\end{equation}
where $K_e = -b/a$ and $T_e = -1/a$ are the engine gain and time constant, respectively. 

Subsequently, a finite set of sample steady-state points, $N_i$ and $W_f(N_i)$, was selected. The steady-state fuel flow was obtained as 
\begin{align}
    W_f(N_i) = -f(N_i)/g(N_i).
\end{align}

At each of these points, the system was assumed to be LTI, and the PI controller was tuned to reach the optimal performance. There are several methods for LTI PID tuning, such as Skogestad's internal model method \cite{bib55}, which accounts for time delays, or the popular yet not always sufficient Ziegler-Nichols methods \cite{bib56}. In this paper, the PI controller was tuned to reach a fast step response with no overshoot, minimizing the integral of quadratic tracking error. To avoid unreasonable gain values, additional regularization of the gains was included. Finally, the controller gains were fitted by 5th-order polynomials. 

To handle varying flight conditions, the gains obtained from the design were treated as corrected gains, $K_{p,corr}$ and $K_{i,corr}$. The true gains, $K_p$ and $K_i$, were computed online as
\begin{align} 
    K_{p} &= K_{p,corr} \bigg(\frac{101325}{p_{1t}}\bigg), \label{eq57} \\ 
    K_{i} &= K_{i,corr} \sqrt{\frac{288}{T_{1t}}} \bigg(\frac{p_{1t}}{101325}\bigg), \label{eq58}
\end{align}
where $p_{1t}$ and $T_{1t}$ were measured in real time.

\subsection{SINDy - IMC} \label{sec5.3}
Since the dead-time effects are neglected in this study, the standard internal model controller (IMC) approach was considered. Details can be found, e.g., in \cite{bib57,bib58}. The derived SINDy model can be utilized for the design of a gain-scheduled IMC controller, which is another proposition of this paper. Assuming the fuel system delay as a first-order lag, the second-order internal model in each steady-state point is given as
\begin{equation} \label{eq59}
    G_i(s) = \frac{K_e(N_i)}{T_p T_e(N_i) s^2+(T_p + T_e(N_i))s +1}.
\end{equation}

The internal model controller is given as 
\begin{equation} \label{eq60}
    G_c(s) = (G_i(s)^-)^{-1} (\tau_fs + 1)^{-2},
\end{equation}
where $G_i^-$ denotes the stable part of the system's transfer function, $G_i$, and $\tau_f$ is a time constant of a filter that renders the controller realizable.

The resulting closed-loop controller is
\begin{equation} \label{eq61}
    G_{IMC}(s) = \frac{G_c(s)}{1 - G_c(s) G_i(s)}.
\end{equation}

Since the system (\ref{eq59}) is inherently stable, and thus $G_i^-=G_i$, the resulting controller is
\begin{align} \label{eq62}
    G_{IMC}(s) &= \frac{G_i^{-1} (\tau_fs +1)^{-2}}{1-G_i^{-1} G_i (\tau_fs +1)^{-2}} \nonumber \\
    &= \frac{G_i^{-1} }{\tau_f^2s^2 + 2 \tau_f s } \nonumber \\
    &= \frac{T_p T_e(N_i) s^2+(T_p + T_e(N_i))s +1}{K_e(\tau_f^2s^2 + 2 \tau_f s)}.
\end{align}

This expression can be rewritten to an alternative form for $\dot{v}$ control. Multiplying both sides by $K_e(\tau_f^2s^2 + 2 \tau_f s)E(s)$, performing an inverse Laplace transform, and integrating both sides yields the following control law:
\begin{align}
    \dot{v} &= \frac{1}{K_e \tau_f^2} \left( T_e T_f \dot{e} +(T_e + T_f)e + \int_0^t e(\tau) d \tau \right) \nonumber \\ &- \frac{2}{\tau_f} v,  \label{eq63} \\
    v &=\int\dot{v} d \tau, \label{eq64}
\end{align}
where the first term on the right-hand side represents a PID controller with the corrected gains given as $K_{p,corr} = (T_e + T_f)/(K_e \tau_f^2)$, $K_{i,corr} = 1/(K_e \tau_f^2)$, and $K_{d,corr} = T_e T_f / (K_e \tau_f^2)$. 

\begin{figure}[!ht] 
    \centering
    \includegraphics[width=1\linewidth]{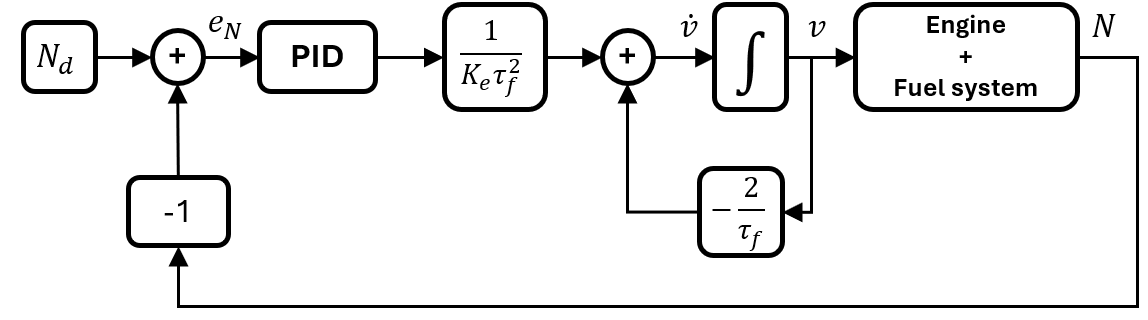}
    \caption{The IMC block diagram}
    \label{fig3}
\end{figure}

The filter time constant $\tau_f$ is the only tuning parameter and can be scheduled as a function of $N$. In this paper, it was assumed to be constant. The higher the $\tau_f$, the lower the sensitivity to measurement noise due to the derivative action as the cut-off frequency decreases, but the slower the response. Thus, the optimal tuning is a trade-off between the noise sensitivity and performance. The corrected parameters are assumed to be as in the previous case. The block diagram of the IMC controller is depicted in Fig. \ref{fig3}.

\subsection{LQI and Kalman observer design} \label{sec5.4}
The LQG controller with integral action (servo LQG or LQGI) can be designed in the eigenfunction space to reach a globally optimal control performance even outside the region of validity of the linearized LPV model. The LQGI design consists of two steps: the LQI controller design and the Kalman observer design. Both steps can be performed individually according to the separation theorem. 

The advantage of designing controllers in the eigenfunction coordinates is that we can employ powerful tools known from the control theory of linear systems and conveniently analyze the stability and robustness. Opposed to the locally linearized systems, the stability and robustness hold not only in local regions around steady-state operating points, but also in the wider operating region of the engine, assuming that this region is sufficiently covered by training data. 

\subsubsection{LQI controller design preliminaries} \label{sec5.4.1}
To design an LQI controller, the eigenfunctions are first augmented with the fuel system dynamics characterized by an FOPDT with time constant $T_f$. The first augmented state vector is denoted as $\boldsymbol{\zeta} = \begin{bmatrix} \mathbf{\Phi} & W_f \end{bmatrix}^\intercal$. Since the functions of input dynamics in the system (\ref{eq53}) are dependent on the spool speed estimate, the system is evaluated at a finite set of evenly distributed sample points, $N_i$, from the $N$ interval. At each point, the system is treated as an LTI system. The fuel system dynamics (\ref{eq10}) in state-space form are augmented with the system, and the resulting matrices are defined as
\begin{equation}  \label{eq65}
    \textbf{A}_\textbf{i}=\begin{bmatrix}
        \mathbf{\Lambda} & \textbf{G}(N_i) \\ \textbf{0} & -\frac{1}{T_f} 
    \end{bmatrix}, \hspace{5mm}
    \textbf{B}_\textbf{i} = 
    \begin{bmatrix}
        \textbf{0} \\ \frac{1}{T_f}
    \end{bmatrix}, \hspace{5mm}
    \textbf{C}_\textbf{i}=\begin{bmatrix}
        \textbf{C} & 0
    \end{bmatrix},
\end{equation}
where $\textbf{G}(N_i)=(\partial \mathbf{\Phi}(N_i)/\partial N) g(N_i)$. 

A new input variable $v$ is introduced. It represents the fuel flow \textit{command} given by the controller. Furthermore, $\boldsymbol{\zeta}$ is augmented with an integral state $\eta$, dynamics of which is given as
\begin{equation} \label{eq66}
    \eta = \int_{t_0}^{t_f} (N_d-N) d\tau, \hspace{5mm} \dot{\eta} = N_d-\textbf{C}\mathbf{\Phi}.
\end{equation}

The final augmented state vector is denoted as $\textbf{z}=\begin{bmatrix}\boldsymbol{\zeta} & \eta \end{bmatrix}^\intercal$, and the resulting state-space model is
\begin{equation}  \label{eq67}
\dot{\textbf{z}} = \mathbf{A_{aug,i} \hspace{0.5mm} z} + \mathbf{B_{aug,i}} \hspace{0.5mm} v + \begin{bmatrix} \textbf{0} \\ 1 \end{bmatrix} N_d, 
\end{equation}
where
\begin{align} \label{eq68}
   \mathbf{A_{aug,i}}=\begin{bmatrix}
    \textbf{A}_\textbf{i} & \textbf{0} \\    
    -\textbf{C}_\textbf{i} & 0
\end{bmatrix}, \quad \mathbf{B_{aug,i}}=\begin{bmatrix}
    \textbf{B}_\textbf{i} \\ 0
\end{bmatrix}. 
\end{align}

The optimal controller feedback gain matrix minimizes the following cost function:
\begin{equation} \label{eq69}
    J_{LQI}=\int_{t_0}^{t_f} \left(\textbf{z}^\intercal\mathbf{Q_z}\textbf{z}+Rv^2\right) d\tau,
\end{equation}
where $\mathbf{Q_z}$ and $R$ are the augmented state and input weighting matrices, respectively.

The final K-LQGI control law is defined for normalized corrected quantities. Thus, the integral state in (\ref{eq66}) must also be computed by first correcting (and normalizing), and then integrating the tracking error. 

This way, the effects of changing flight conditions are captured to some extent. However, for higher Mach numbers of flight, the corrections start losing validity due to compressibility effects.  

\subsubsection{Optimization algorithm} \label{sec5.4.2}
MATLAB provides a simple way to solve the LQI problem (\ref{eq69}) using the Riccati equation. Another task is to design the weighting matrices such that the overall performance is optimal. For this purpose, the metaheuristic approach, e.g., genetic algorithm (GA) or PSO, can be utilized. The algorithm minimizes a custom cost function based on the integral of absolute error (IAE). Furthermore, since the specified stability margins must be respected, the problem is constrained. Thus, GA and PSO are suitable tools \cite{bib53,bib59}. 

The global controller cost function is a sum of a weighted IAE and additional terms accounting for the maximum overshoot and undershoot of the maximum value, $\bar{N}_{max}=1$, and the minimum value, $\bar{N}_{min} = 0.5$, respectively, where $\bar{N}$ denotes the physical spool speed relativized using the nominal value of 14000 RPM ($\bar{N} = N_{phys}/14000$). These terms also penalize not reaching the maximum and minimum commands fast enough. The weighting by $w_e$ is performed so that the error during smaller transients has the same magnitude as during the large transients to even out their importance. 

The controller optimization problem is given as
\begin{align} \label{eq71}
    \min_{\mathbf{Q_z}, R} &\sum_{k=1}^{n_s-1} w_e\left(\frac{|e_{k}| + |e_{k+1}|}{2}\right)\Delta t \nonumber  \\ &+ \alpha_{os} |\max(\bar{N}) - 1| \nonumber \\
    &+ \alpha_{us} |\min(\bar{N})-0.5|,
\end{align}
s.t.
\begin{align}
  & e_{k} = \bar{N}_d(t_k) - \bar{N}(t_k), \label{eq72} \\
  & \bar{N}(t_k) = (10000N(t_k) + 5000)/14000, \\
  & N(t_k) = \mathbf{C \Phi} (t_k), \label{eq77} \\
  & \mathbf{\dot{\Phi} = \Lambda\Phi + G}(N) W_f, \label{eq76} \\
  & W_{f,k+1} = \left(1-\Delta t/T_f\right)W_{f,k}+\left(\Delta t/T_f\right) v_k , \label{eq75}\\
  & v_{min} \le v \le v_{max} , \label{eq73} \\
  & \dot{\eta} = 0 \quad \text{when} \quad v=v_{min} \lor v=v_{max},\label{eq74} \\
    &w_e = \begin{cases}
       5 &\quad\text{for } 5 \le t < 50 \\
       0 &\quad\text{for } t < 5 \text{, and } 50 \le t < 55, \label{eq78} \\
       1 &\quad\text{for } 55 \le t \le 70
     \end{cases} \\
     & GM_{min} > 6\text{ dB}, \label{eq79} \\
     & PM_{min} > 60 \text{ deg}, \label{eq80}
\end{align}
where $\alpha_{os}$ and $\alpha_{us}$ are the overshooting and undershooting weights, respectively, $n_s$ is the number of simulation time steps, and $GM_{min}$ and $PM_{min}$ are the minimum open-loop gain and phase margins based on the entire range of $N_i$, respectively.

\begin{figure}[!ht] 
    \centering
    \includegraphics[width=1\linewidth]{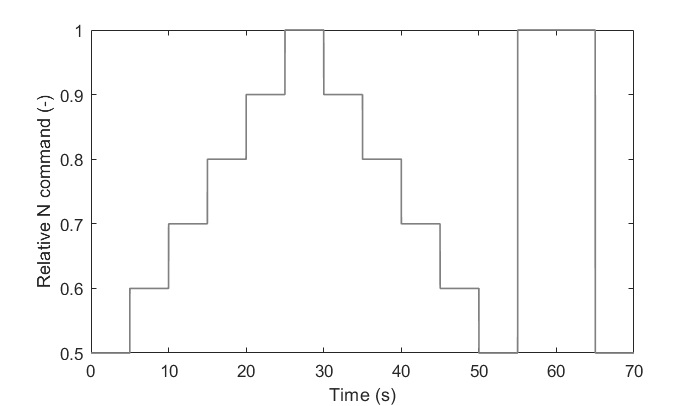}
    \caption{The relative spool speed command profile utilized for the control law optimization}
    \label{fig4}
\end{figure}

Numerical integration is utilized to simulate the system's response to a predefined testing spool-speed command profile depicted in Fig. \ref{fig4}. This profile captures both the small and large transients. The response is used for cost calculation according to (\ref{eq71}).

Among the optimized parameters are the weight matrix of the transient spool speed $\mathbf{Q_N}$, integral action $\mathbf{Q_i}$, delayed fuel flow $\mathbf{Q_f}$, and input $\mathbf{R_c} = R$. These are further complemented by a weight perturbation diagonal matrix $\mathbf{dQ_\Phi}$ targeting individual spectral components. This allows the algorithm to balance a possibly more aggressive integral action. The complete state weighting matrix is given as
\begin{equation} \label{eq81}
    \mathbf{Q_z} = 
    \begin{bmatrix}
        \mathbf{Q_\Phi + dQ_\Phi} & \textbf{0} & \textbf{0}\\
        \textbf{0} & \textbf{Q}_\textbf{f} & \textbf{0} \\
        \textbf{0} & \textbf{0} & \textbf{Q}_\textbf{i}
    \end{bmatrix},
\end{equation}
where the $\mathbf{Q_\Phi}$ matrix is generally obtained as
\begin{align} \label{eq82}
    \mathbf{x^\intercal Q_x x} &= \mathbf{(C\Phi)^\intercal Q_x C \Phi} \nonumber \\
    &= \mathbf{\Phi^\intercal C^\intercal Q_x C \Phi} \nonumber \\
    &= \mathbf{\Phi^\intercal Q_\Phi \Phi}.
\end{align}
Thus, for the GTE, it yields $\mathbf{Q_\Phi = C^\intercal Q_N C}$. 

The input $v$ follows a closed-loop LQI control law, which will be presented in the following paragraphs. 

As before, the control law is designed for a finite set of sample points $N_i$ evenly distributed across the normalization interval $[N_{min}, N_{max}]$. At each point, the system is considered LTI with a controllable pair $(\mathbf{A_{aug,i}, B_{aug,i}})$, and the following equations are solved:
\begin{equation}    \label{eq83}
\mathbf{K_{LQI,i} = R_c^{-1} B_{aug,i}^\intercal P_{c,i}},
\end{equation}
where $\mathbf{P_{c,i}}$ is a corresponding solution of the CARE

\begin{align} \label{eq84}
\mathbf{A_{aug,i}^\intercal P_{c,i}} & \mathbf{+ P_{c,i} A_{aug,i} + Q_z} \nonumber \\
    & \mathbf{- P_{c,i} B_{aug,i} R_c^{-1} B_{aug,i}^\intercal P_{c,i} = 0},
\end{align}
where $\mathbf{R_c}$ is the controller input weighting matrix.

Subsequently, the feedback gain matrix entries are interpolated or approximated by a suitable function of $N$, and the resulting gain-scheduled LQI control law is given as  
\begin{equation} \label{eq85}
    v = -\mathbf{K_{LQI}}(\hat{N}) \hspace{0.5mm}  \begin{bmatrix} \boldsymbol{\zeta} & \eta \end{bmatrix} ^\intercal.
\end{equation}

\subsubsection{Kalman observer design} \label{sec5.4.3}
To improve the prediction performance of the baseline KEM, a Kalman observer can be designed. Introducing noise, the KEM equations are given as
\begin{align}\label{eq85-2}
    &\mathbf{\dot{\Phi} = \Lambda\Phi + \nabla\Phi} g u+\mathbf{w_p}, \quad \mathbf{w_p} \sim \mathcal{N}(\mathbf{0,\Sigma_p}), \\ \nonumber
    &\hat{N} = \mathbf{C\Phi} + w_n, \quad w_n \sim \mathcal{N}(0,\sigma_n^2), 
\end{align}
where $\mathbf{w_p}$ and $w_n$ are the process and measurement noise, respectively, both assumed to be Gaussian white noise with covariances $\mathbf{\Sigma_p}$ and $\sigma_n^2$, respectively. 

\begin{figure}[!ht] 
    \centering
    \includegraphics[width=0.9\linewidth]{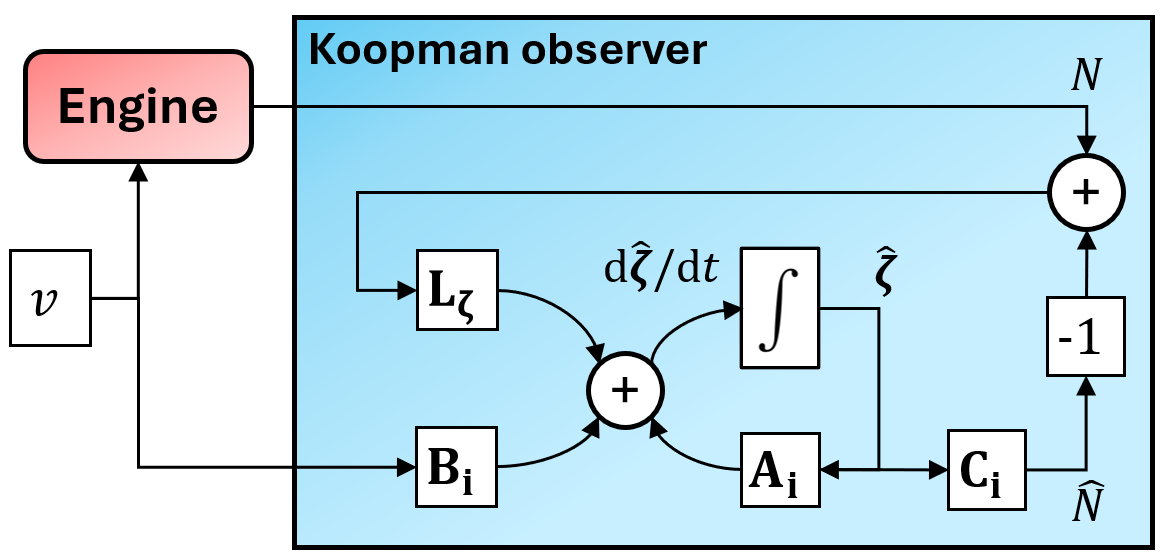}
    \caption{The Kalman observer block diagram}
    \label{fig5}
\end{figure}

Since both $\mathbf{\Lambda}$ and $\textbf{C}$ matrices are time-invariant, the steady Kalman filter can be obtained as a solution to the estimator CARE. This was discussed in, e.g., \cite{bib60}. The observer state-space model is shown in Fig. \ref{fig5}. 

The resulting state equation is given as
\begin{align} \label{eq86}
    \dot{\hat{\boldsymbol{\zeta}}} &= \mathbf{A_i}\hat{\boldsymbol{\zeta}} + \mathbf{B_i}v + \textbf{L}_{\boldsymbol{\zeta}} (N-\mathbf{C_i}\hat{\boldsymbol{\zeta}}) \nonumber \\
    &=(\mathbf{A_i} - \textbf{L}_{\boldsymbol{\zeta}} \mathbf{C_i})\hat{\boldsymbol{\zeta}}+\mathbf{B_i} v+\textbf{L}_{\boldsymbol{\zeta}} N,
\end{align}
where the Kalman gain matrix $\textbf{L}_{\boldsymbol{\zeta}}$ is constant. 

As can be seen, the fuel flow state $W_f$ is also included, since the equation is formulated for $\boldsymbol{\zeta}$. The gain matrix can be further written as $\textbf{L}_{\boldsymbol{\zeta}} = \begin{bmatrix} \textbf{L} & 0 \end{bmatrix}^\intercal$, where, during the design phase, the matrix $\textbf{L}$ is optimized only for $\mathbf{\Phi}$, yielding a static Kalman observer, and the fuel flow estimation depends only on the first-order model. Thus, $\hat{W}_f = W_f$ as demonstrated in the following equation:
\begin{equation} \label{eq87}
    \hat{W}_{f,k+1}=W_{f,k+1}=\left(1-\frac{\Delta t}{T_f}\right)\hat{W}_{f,k}+\frac{\Delta t}{T_f}v_k.
\end{equation}

Thus, the $W_f$ dynamics are accounted for only during the LQI design. This assumption was deemed valid since the first-order model of the fuel system was considered accurate enough to provide reliable information about the fuel flow with no need for Kalman correction.  

If the estimation of $W_f$ is still of interest, it would be more advantageous to switch to the discrete-time formulation of the KEM and utilize the time-varying Kalman filter, as the system matrix would become time-varying. 

Several approaches can be used for the observer design, see, e.g., \cite{bib60,bib61,bib62}. In this paper, similarly to the LQI design step, assuming that the pair $(\mathbf{\Lambda,C})$ is observable, the CARE approach is considered, and the Kalman gain is given by the following equations:s
\begin{equation} \label{eq88}
    \mathbf{L=P_o C^\intercal R_o^{-1}},
\end{equation}
where $\textbf{P}_\textbf{o}$ is a solution of the observer CARE:

\begin{equation} \label{eq89}
    \mathbf{\Lambda P_o + P_o \Lambda^\intercal +G_o Q_o G_o^\intercal - P_o C^\intercal R_o^{-1} C P_o = 0},
\end{equation}
where $\mathbf{G_o}$ is the process noise gain matrix usually set to $\textbf{I}$, $\mathbf{R_o}$ is the measurement noise covariance matrix estimated from data, and $\mathbf{Q_o}$ is the process noise covariance matrix. The latter can be estimated using the test dataset eigenfunction dynamics error covariance $\text{Cov}(\Delta\mathbf{\Phi}(t) - \Delta\mathbf{\Phi}(N(t)))/\Delta t$, where $\mathbf{\Phi}(N(t))$ are eigenfunctions evaluated using the ground-truth data. Alternatively, $\mathbf{Q_o}$ can also be estimated from the eigenfunction time derivative error covariance since the process noise $\mathbf{w_p}$ is acting on the dynamics in (\ref{eq85-2}). Thus, the process noise covariance estimate is given as
\begin{align} \label{eqCov}
    \mathbf{\Sigma_p} \approx \text{Cov} \bigg(\frac{\partial \mathbf{\Phi}(N(t))}{\partial N} \dot{N} & -\mathbf{\Lambda\Phi}(t) \nonumber \\
    &- \nabla \mathbf{\Phi}\big(\hat{N}(t)\big) g\big(\hat{N}(t)\big) u(t) \bigg).
\end{align}

The resulting Koopman LQGI control law is given using the estimated state vector as
\begin{equation} \label{eq90}
    v= -\mathbf{K_{\boldsymbol{\zeta}}}(\hat{N})  \hat{\boldsymbol{\zeta}} - K_i(\hat{N}) \eta,
\end{equation}
and $\mathbf{K_{LQI}} = \begin{bmatrix} \textbf{K}_{\boldsymbol{\zeta}} &  K_i\end{bmatrix}$.

Inserting the control law in (\ref{eq90}) to (\ref{eq67}) and augmenting with the Kalman filter, (\ref{eq86}), according to the separation theorem \cite{bib62}, we obtain the following closed-loop state-space model:
\begin{align} \label{eq91}
    \begin{bmatrix}
        \dot{\boldsymbol{\zeta}} \\ \dot{\boldsymbol{\varepsilon}} \\ \dot{\eta}
    \end{bmatrix} &= 
    \begin{bmatrix}
        \mathbf{A_i - B_i K_{\boldsymbol{\zeta}}}  & \mathbf{B_i K_{\boldsymbol{\zeta}}} &-\mathbf{B_i} K_i \\
        \textbf{0} & \mathbf{A_i - L_{\boldsymbol{\zeta}} C_i} & \textbf{0} \\
        -\mathbf{C_i} & \textbf{0} & 0
    \end{bmatrix}
     \begin{bmatrix}
        \boldsymbol{\zeta} \\ \boldsymbol{\varepsilon} \\\eta
    \end{bmatrix} \nonumber \\ &+ 
     \begin{bmatrix}
        \textbf{0} \\ \textbf{0} \\ 1
    \end{bmatrix} N_d,
\end{align}

\begin{equation} \label{eq92}
    \hat{N} = \textbf{C}_\textbf{i} \hat{\boldsymbol{\zeta}} = \textbf{C}_\textbf{i} \boldsymbol{(\zeta-\varepsilon)} = \begin{bmatrix}
        \textbf{C}_\textbf{i } & -\textbf{C}_\textbf{i}  & 0
    \end{bmatrix} 
    \begin{bmatrix}
        \boldsymbol{\zeta} \\ \boldsymbol{\varepsilon} \\\eta
    \end{bmatrix},
\end{equation}
where $\hat{\boldsymbol{\zeta}}=\begin{bmatrix}
    \hat{\mathbf{\Phi}} & \hat{W}_f
\end{bmatrix}^\intercal$ and $\boldsymbol{\varepsilon=\zeta-\hat{\zeta}}=\begin{bmatrix}\mathbf{\Phi-\hat{\Phi}} & 0\end{bmatrix}^\intercal$ are the augmented state and estimation error vectors, respectively. 

In the open-loop model, the $-\mathbf{C_i}$ matrix in the second row is replaced by $\textbf{0}$, as the integral of the tracking error changes to the integral of the demanded output only. Note that $\eta$ is computed using the real measured $N$, not $\hat{N}$. Given the noisy spool-speed measurement, the controller gains and KEM input dynamics are evaluated using the Kalman estimate, $\hat{N}$.

\begin{figure*}[!ht] 
    \centering
    \includegraphics[width=0.75\linewidth]{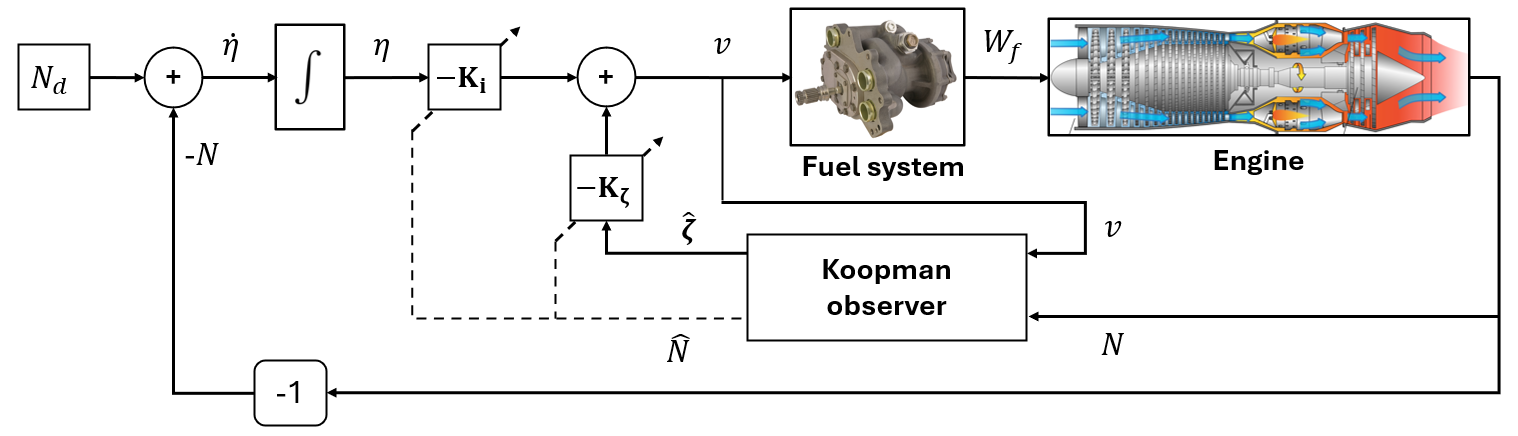}
    \caption{The entire control system with KEM-LQGI controller. Solid arrows represent the main signal flow, and dashed arrows depict the scheduling signals}
    \label{fig6}
\end{figure*}

The complete control system is depicted in Fig. \ref{fig6}

\section{Results}\label{sec6}

\subsection{Discovery of turbojet governing equation from data} \label{sec6.1}

\subsubsection{Data acquisition and processing} \label{sec6.1.1}
The training and test datasets were obtained from simulations using the in-house nonlinear CLM. Validation of this model against the commercial simulation software GasTurb was performed in \cite{bib40}. The training dataset consisted of three 500-second segments. The first segment was formed by faster ramps with a rise time in the range of 1-2 seconds. The command was changed approximately every 10 seconds. The second segment was a chirp signal with a frequency sweep from 0 to 0.25 Hz, aiming to properly capture the low-frequency response. The last segment consisted of slow ramps with longer duration in the steady state, as the model must learn the steady-state regions well to avoid the prediction drifting away. The goal was to capture various aspects of the GTE dynamics. The test dataset was designed similarly, but with overall faster dynamics. The sampling period was 10 ms for all simulations.

\begin{figure}[h] 
    \centering
    \begin{subfigure}[h]{0.4\textwidth}
        \centering
        \includegraphics[width=1\linewidth]{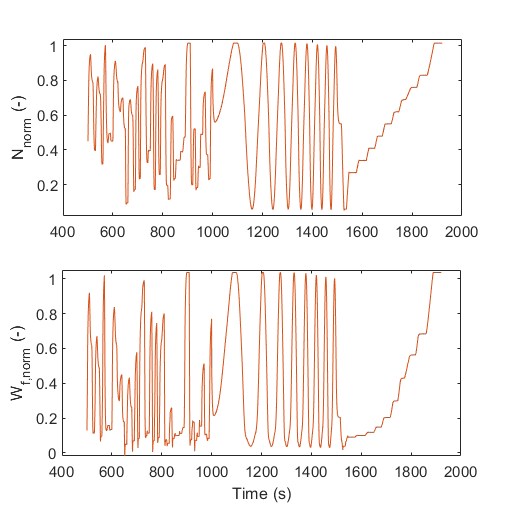}
        \caption{}
        \label{fig7a}
    \end{subfigure}
    \hspace{5mm}
    \begin{subfigure}[h]{0.4\textwidth}
        \centering
        \includegraphics[width=1\linewidth]{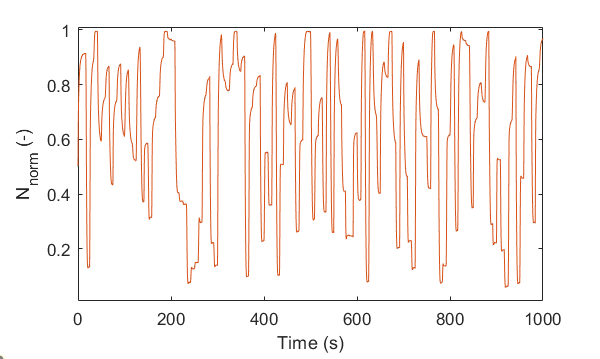}
        \caption{}
        \label{fig7b}
    \end{subfigure}
    \caption{The normalized (a) training and (b) test datasets}
    \label{fig7}
\end{figure}

To simulate the measured data, the smooth simulation output was further superimposed with Gaussian white noise with a standard deviation of 50 RPM generated using the MATLAB \textit{randn} command. This slightly exceeds the typical RPM-sensor uncertainty standard deviation reported to be $0.25$ $\%$ of nominal RPM for gas turbines \cite{Wei2025}. Furthermore, filtering of the datasets was performed using a Savitzky-Golay polynomial filter. After subsequent normalization, the time derivatives of normalized spool speed were calculated using a 2nd-order finite-difference numerical differentiation scheme. The training and test datasets are depicted in Fig. \ref{fig7a} and Fig. \ref{fig7b}, respectively. 

\subsubsection{SINDy} \label{sec6.1.2}
For identification of the GTE's governing equation, the SINDy algorithm with logistic functions (LFs) was utilized. LFs are a popular choice for activation functions in neural networks, and they have favorable properties \cite{bib63}. As explained in Section \ref{sec4.1}, SINDy was combined with GD and ADAM optimizer and subjected to the L1 regularization of linear coefficients. Additionally, specifically for the LFs and basically for any RBF, the sparsity could be further achieved through the nonlinear parameters. Thus, these were optimized as well instead of just being guessed, or assumed to be evenly distributed, as is a common practice. The equation of LF is given as
\begin{equation} \label{eq93}
    LF(x) = \frac{1}{1+e^{-\varepsilon(x-\mu)}},
\end{equation}
where $\varepsilon$ and $\mu$ are the steepness parameter and center point, respectively.

The following properties of LFs can help exclude some inactive or constant terms from the SINDy model: 
\begin{itemize}
    \item $\xi\rightarrow 0$ for inactive terms via regularization. 
    \item $\mu\rightarrow\infty\ \cup\ \varepsilon>0$ or $\mu\rightarrow-\infty\ \cup\ \varepsilon<0$ causes LF to be 0.
    \item $\mu\rightarrow-\infty \cup \varepsilon>0$ or $\mu\rightarrow\infty\ \cup \varepsilon<0$ causes LF to be 1 (constant).
    \item $\varepsilon=0$ causes LB to be 0.5 (constant). 
\end{itemize}

In the latter two cases, constants can be summed up to form a single constant function. The L1 regularization parameter was set to $10^{-4}$, and the zero derivative condition parameter was set to 0.005. The initial learning rates were set as 0.01, 50, and 0.01 for $\Xi$, $\varepsilon$, and $\mu$, respectively. The algorithm minimized a cost function given by the static $\dot{N}$ data. Additionally, every 10 iterations, the differential equation (\ref{eq23}) was solved numerically for validation, and the true prediction MAE was stored. Once the prediction MAE started to grow, e.g., due to overfitting, the algorithm stopped.

\begin{figure}[!ht] 
    \centering
    \includegraphics[width=1.1 \linewidth]{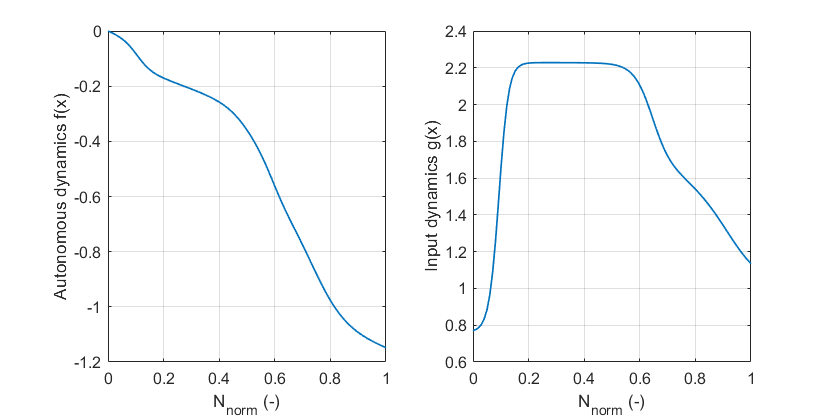}
    \caption{Autonomous and input dynamics functions obtained from SINDy}
    \label{fig8}
\end{figure}

\begin{figure}[!ht] 
    \centering
    \includegraphics[width=1\linewidth]{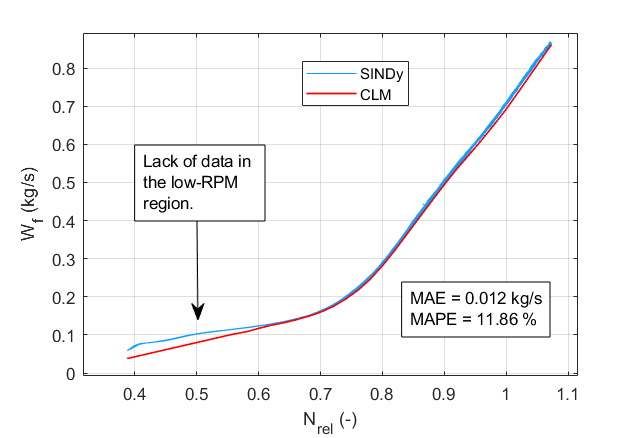}
    \caption{Comparison of steady-state fuel flow characteristics}
    \label{fig9}
\end{figure}

The SINDy was initially performed for 5 LFs, both for autonomous and input dynamics. Additional terms were given as $N, NW_f, W_f$, yielding a total of 13 terms. After the first round of the GD, the first LF in $f(N)$, and the $NW_f$ term were excluded, and the rest of the terms were updated in the second round with a smaller regularization parameter of $10^{-5}$. The resulting SINDy functions are shown in Fig. \ref{fig8}. It can be seen that the condition $f(0)=0$ was satisfied. Additionally, the steady-state fuel-flow characteristics were compared as shown in Fig. \ref{fig9}. For the SINDy model, the fuel flow was computed as $W_f=-f(N)/g(N)$ and de-normalized.

\begin{figure}[!ht] 
    \centering
    \includegraphics[width=1\linewidth]{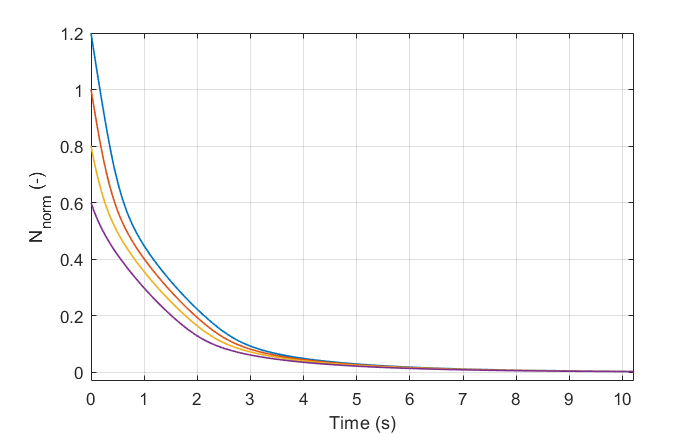}
    \caption{Autonomous trajectories of the GTE from the estimated $f(N)$ function} 
    \label{fig10}
\end{figure}

Subsequently, the autonomous dynamics were extracted, and trajectories starting from multiple initial conditions were generated as plotted in Fig. \ref{fig10}.

\subsection{KEM construction from the SINDy model and the Kalman observer design} \label{sec6.2}
Based on the generated trajectories, the PSO was utilized to solve the constrained minimization problem (\ref{eq36}) and to discover the optimal set of eigenvalues. Both the real and complex eigenvalues were considered and compared. The minimum and maximum real parts were set as -10 and -0.3, respectively; the maximum imaginary part was 10, and the maximum $\beta/\alpha$ ratio was 2.5. The PSO population size was 200, and the velocity weights were kept default. 

Regarding the real eigenvalues, the automatic switching from distinct to repeated when necessary was enabled, and the repeated eigenvalues with secular terms led to mean absolute error (MAE) of $0.001$. However, the complex eigenvalues outperformed the repeated eigenvalues with MAE of $4.5 \times 10^{-4}$, and thus were utilized for the controller design. The system order was gradually increased from 4 to 8, and it was found that the 6th-order is an optimal trade-off between the complexity and accuracy.  

\begin{figure}[!ht] 
    \centering
    \includegraphics[width=1\linewidth]{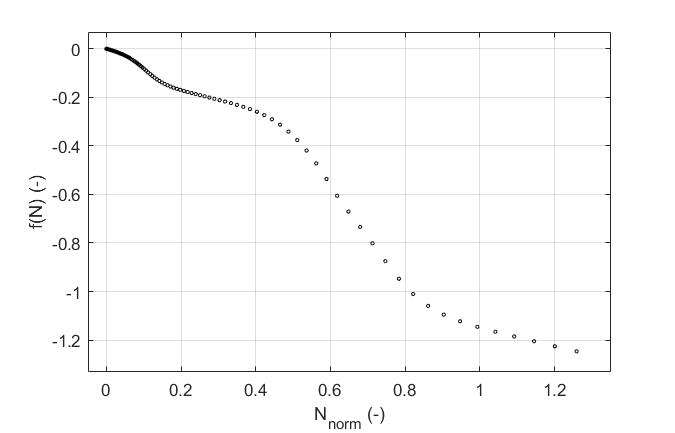}
    \caption{Sampled autonomous function with 200 sample points}
    \label{fig11}
\end{figure}

After the eigenvalue optimization, the GD algorithm with the trivial solution avoidance was employed to solve the eigenfunction KPDEs given by Eqs. (\ref{eq48}) and (\ref{eq49}). For the $f(N)$ sampling, the rather aggressive exponential sampling from Fig. \ref{fig2} was considered with 500 samples. The sampled function $f(N)$ is plotted in Fig. \ref{fig11} for only 200 samples for clarity. 

Originally, a set of 20 LFs was optimized for each eigenfunction. However, it was later found out that just 10 LFs were also sufficient. This reveals yet another advantage over DMD-based approaches – as the basis functions can adapt to each eigenfunction individually, eigenfunctions can have a different number of terms, depending on the complexity of the eigenfunction's shape. Furthermore, the basis can be decided based on the discovered shapes, avoiding the trial-and-error selection for EDMD. 

\begin{figure}[!ht] 
    \centering    
    \includegraphics[width=1\linewidth]{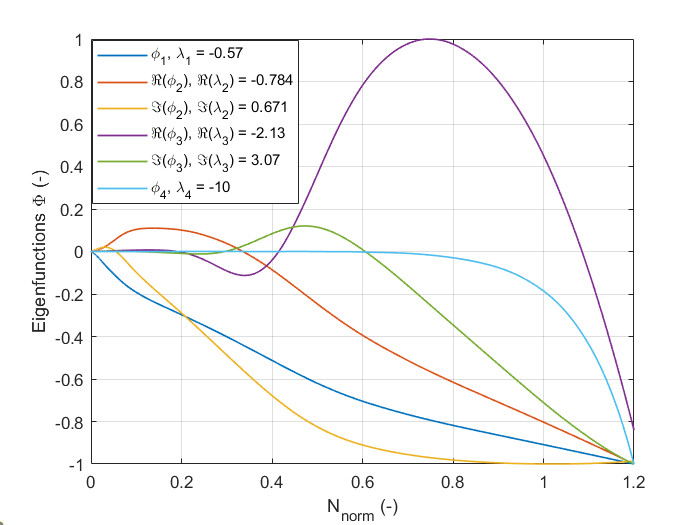}
    \caption{The resulting eigenfunctions}
    \label{fig12}
\end{figure}

The resulting eigenfunctions are shown in Fig. \ref{fig12}. Subsequently, the vector of mode amplitudes was found using (\ref{eq51}) (or (\ref{eq52})). 

\begin{figure}[!ht]
    \centering
    \includegraphics[width=1\linewidth]{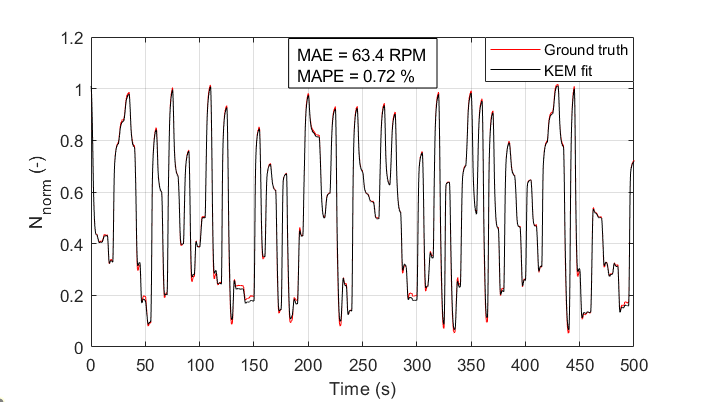}
    \caption{Test dataset performance for the first 500 seconds}
    \label{fig13}
\end{figure}

Following the identification, the final model was validated using the test dataset. The validation was performed for the filtered data to assess the “off-line” prediction accuracy, which is not affected by noise. The MAE and the mean absolute percentage error (MAPE) were utilized as the error metrics. They were computed for the de-normalized corrected spool speed. For the training data, the final MAE and MAPE were 28.44 RPM and 0.33 $\%$, respectively. The test dataset prediction results, MAE, and MAPE are depicted in Fig. \ref{fig13}. 

\begin{figure}[!ht] 
    \centering 
    \begin{subfigure}[h]{0.4\textwidth}
        \centering
        \includegraphics[width=1.2\linewidth]{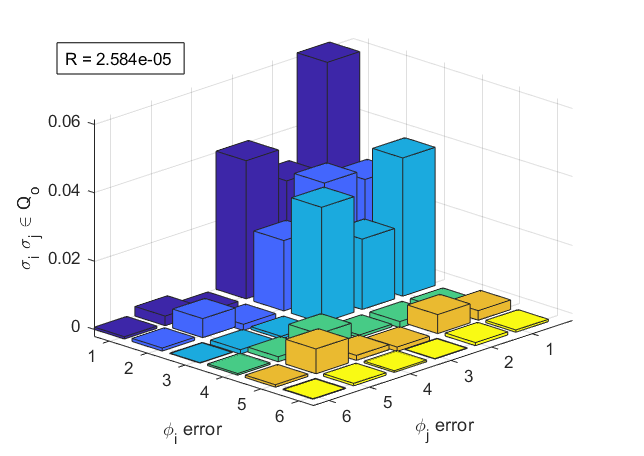}
        \caption{}
        \label{fig14a}
    \end{subfigure}
    \hspace{5mm}
    \begin{subfigure}[h]{0.4\textwidth}
        \centering
        \includegraphics[width=1.2\linewidth]{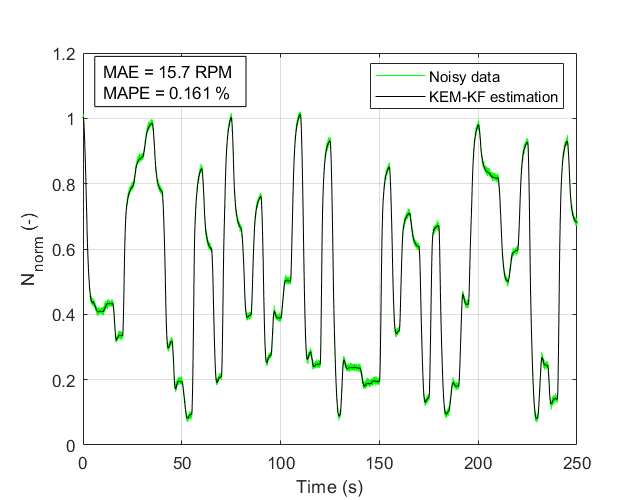}
        \caption{}
        \label{fig14b}
    \end{subfigure}
    \caption{(a) Eigenfunction prediction error covariance matrix bar plot, and (b) the Kalman filter/observer estimate against the noisy data}
    \label{fig14}
\end{figure}

Subsequently, the Kalman observer was designed. The covariance matrix of the eigenfunction prediction error was computed and is depicted in Fig. \ref{fig14a}. It can be seen that there are off-diagonal terms that are not negligible. This matrix was utilized for tuning of the process noise weighting matrix $\mathbf{Q_o}$. The measurement noise covariance $\mathbf{R_o}$ was estimated from a steady-state sample taken from unfiltered normalized spool-speed data, and its value is also shown in the figure. The performance of KEM with the Kalman observer with MAE and MAPE is depicted in Fig. \ref{fig14b}. Both metrics were calculated based on the difference between the estimation and clean data (with no noise). It can be concluded that the Kalman observer provides very accurate estimation of both the eigenfunctions (directly) and also the spool speed (indirectly). 

\begin{figure}[!ht]
    \centering
    \includegraphics[width=1\linewidth]{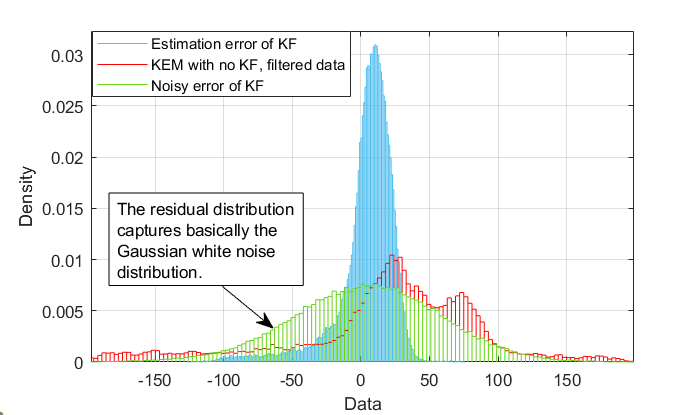}
    \caption{Residual histograms of the baseline KEM without the KF for the dataset filtered using the Savitzky-Golay filter (red), error of prediction of the noisy dataset using the KF that reveals the Gaussian white noise distribution with variance of 50 RPM (green), and estimation error computed using clean data without noise (blue)}
    \label{fig15}
\end{figure}

Figure \ref{fig15} depicts the residual histograms. These are evaluated for a) the prediction using the baseline KEM model for the test dataset filtered using the Savitzky-Golay filter (red), b) the prediction of the noisy test dataset using the Kalman observer, which reveals the Gaussian distribution of the added white noise (green), and c) residuals given by the difference between the estimation of the Kalman observer and the clean test dataset (with no added noise) to assess the estimation performance (blue). 

\subsubsection{Thrust estimation} \label{sec6.2.1}
As the direct thrust control also belongs to modern research topics of high interest, the normalized corrected thrust from another test dataset was projected onto the span of $\hat{\boldsymbol{\zeta}}(t)$. The thrust output equation and the corresponding LS problem are given as
\begin{align}
&\hat{F} = \mathbf{C_F \hat{\Phi}} + \mathbf{D_F} W_f, \label{eq94} \\   
    \begin{bmatrix}\mathbf{C_F} & \mathbf{D_F}\end{bmatrix} = \arg &\min_\mathbf{C_F,D_F} \| F_{c,n} - \mathbf{C_F}  \mathbf{\Phi} - \mathbf{D_F}W_f \|_2^2, \label{eq95}
\end{align}
where $\mathbf{C_F}$ and $\mathbf{D_F}$ are vectors of coefficients of projection onto $\mathbf{\hat{\Phi}}$ and $W_f$, respectively, and $F_{c,n}$ is the normalized corrected thrust given as
\begin{equation} \label{eq96}
    F_{c,n} = F \frac{101325}{p_{1t}} \frac{1}{F_{DP}},
\end{equation}
where $F_{DP}$ is the design-point thrust. 

\begin{figure}[!ht] 
    \centering
    \includegraphics[width=1\linewidth]{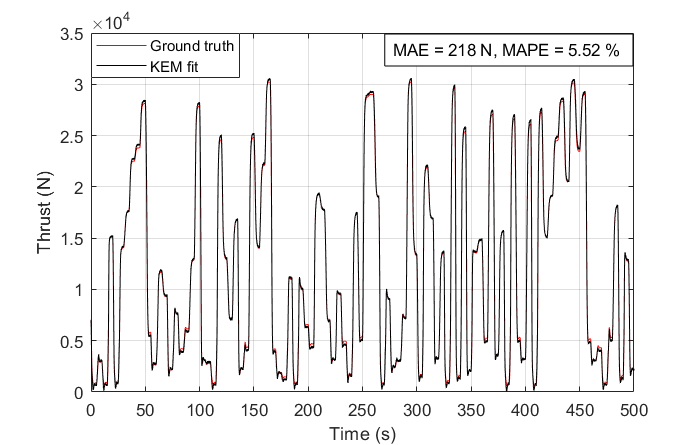}
    \caption{The thrust estimation results for the first 500 seconds of another test dataset}
    \label{fig16}
\end{figure}

The thrust prediction results are shown in Fig. \ref{fig16}. The MAPE is inherently higher because the lower thrust boundary is near 0, and thus the relative error grows exponentially for the same MAE. The results demonstrate that the KEM can also be effectively utilized for thrust control. In this paper, however, only the control of the spool speed is considered.

\subsection{Controller design} \label{sec6.3}

\subsubsection{Classical and gain scheduled nonlinear PI controllers} \label{sec6.3.1}
Next, the controllers were designed. The classical PI controller was tuned to satisfy the aforementioned requirements arising from the PART 31 regulation, stating that the engine should be able to respond to a 1-second linear ramp command from 15 $\%$ to 95 $\%$ of the maximum thrust in under 5 seconds. The resulting gains were given as $K_p = 1$ and $K_i = 3$.

\begin{figure}[!ht] 
    \centering
    \includegraphics[width=1\linewidth]{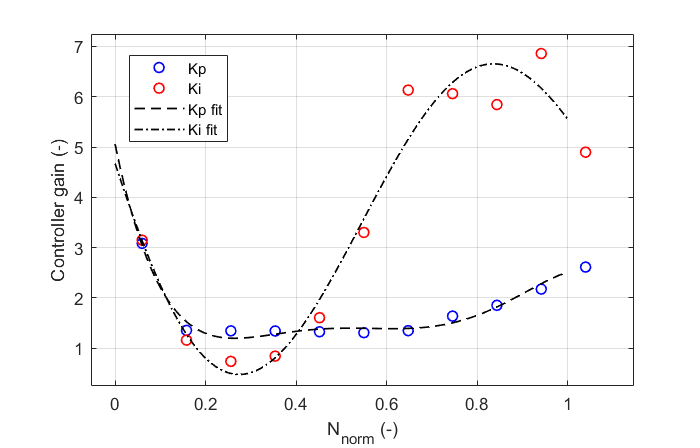}
    \caption{The optimized controller gains with the corresponding fitted polynomials}
    \label{fig17}
\end{figure}

For the LPV-PI design, the discovered SINDy model was analytically differentiated, and the engine gain and time constant were assessed for a set of 11 evenly distributed steady-state points. At each steady-state point, the PI controller was tuned to reach a fast step response with no overshoot, minimizing the integral of quadratic tracking error. To avoid unreasonable gain values, additional regularization of the gains was included. The optimal PI gains were fitted by 5th-order polynomials as depicted in Fig. \ref{fig17}.

\subsubsection{SINDy - IMC} \label{sec6.3.2}
\begin{figure}[!ht] 
    \centering
    \includegraphics[width=1\linewidth]{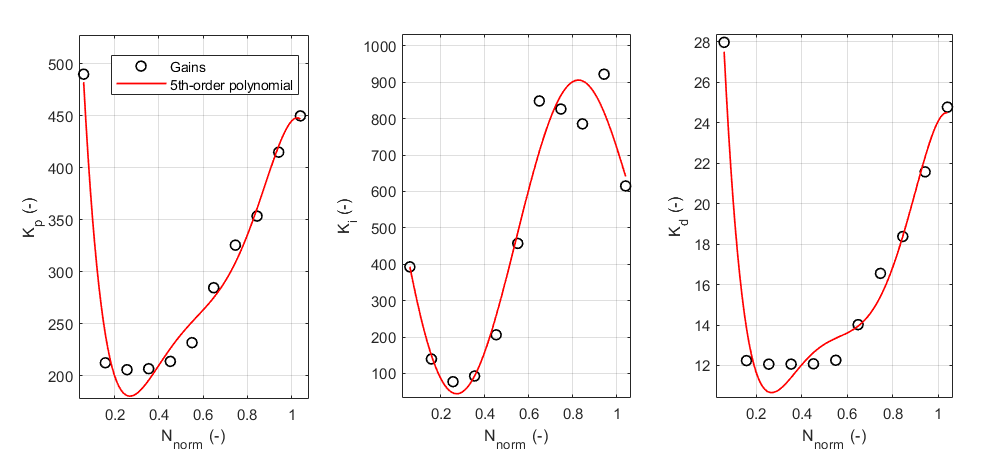}
    \caption{Resulting PID gains of the SINDy-IMC controller}
    \label{fig18}
\end{figure}

The filter time constant was manually tuned to $\tau_f = 0.05$ s. Lower values significantly increased the sensitivity to the measurement noise, whereas higher values worsened the tracking performance. The resulting controller gains of the PID part are plotted in Fig. \ref{fig18}. It can be observed that the trends of proportional and integral gains are similar to those of the LPV-PI controller. 

\subsubsection{Koopman LQGI controller} \label{sec6.3.3}
\begin{figure}[!ht] 
    \centering
    \begin{subfigure}{0.4\textwidth}
        \centering
        \includegraphics[width=1\linewidth]{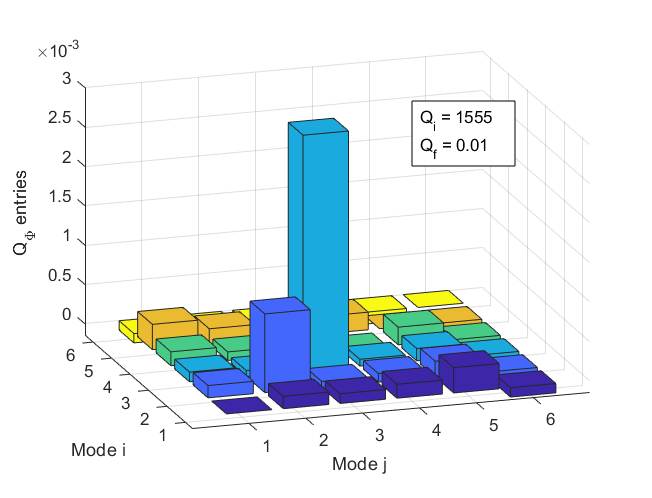}
        \caption{}
        \label{fig19a}
    \end{subfigure}
    \hspace{2mm}
    \begin{subfigure}{0.4\textwidth}
        \centering
        \includegraphics[width=1\linewidth]{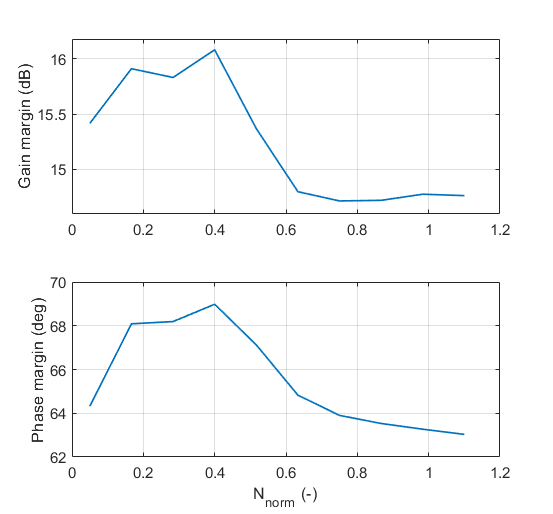}
        \caption{}
        \label{fig19b}
    \end{subfigure}
    \caption{(a) The optimized $\mathbf{Q_\varphi}$ matrix, and (b) the K-LQGI classical open-loop gain and phase margins as functions of the normalized spool speed}

\end{figure}

\begin{figure}[!ht] 
    \centering
    \begin{subfigure}{0.4\textwidth}
        \centering
        \includegraphics[width=1\linewidth]{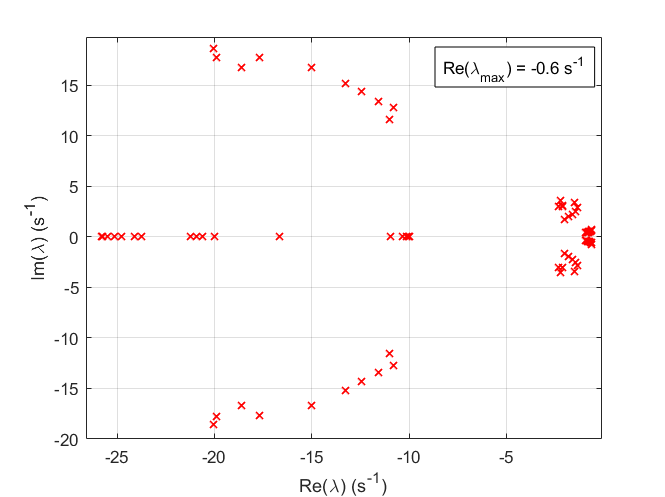}
        \caption{}
        \label{fig20a}
    \end{subfigure}
    \hspace{2mm}
    \begin{subfigure}{0.4\textwidth}
        \centering
        \includegraphics[width=1\linewidth]{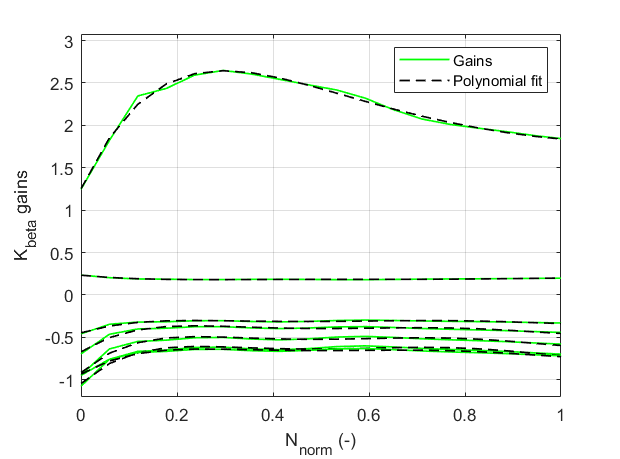}
        \caption{}
        \label{fig20b}
    \end{subfigure}
    \caption{(a) The range of eigenvalues of the closed-loop system with changing spool speed and (b) the $\mathbf{K_{\boldsymbol{\zeta}}}$ gains}
    \label{fig20}
\end{figure}

\begin{figure}[!ht] 
    \centering
    \includegraphics[width=1\linewidth]{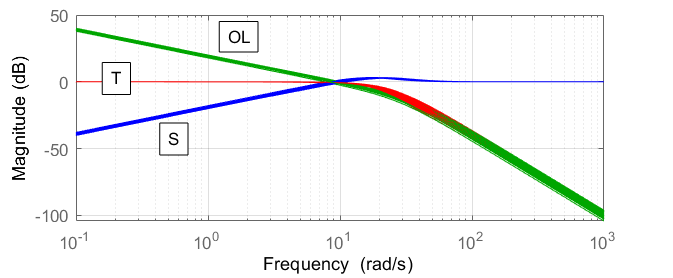}
    \caption{Magnitude Bode plot of open-loop (OL), sensitivity (S), and complementary sensitivity (T) transfer functions}
    \label{fig21}
\end{figure}

The LQGI controller was designed using the proposed approach with GA or PSO from the MATLAB Global Optimization Toolbox. GA yielded better results after multiple initializations. Regarding the hyperparameters, the population size and number of generations were both set to 100, the elite count was set to 3, and the replication-crossover-mutation rates and parameters were kept at the default. The initial population was generated using a Latin hypercube sampling function. The resulting optimal $\mathbf{Q_\Phi}$ matrix is plotted in Fig. \ref{fig19a}. The $\mathbf{Q_f}$ and $\mathbf{Q_i}$ values are also shown, and it can be seen that the algorithm generated a rather aggressive integral gain, possibly balanced by the $\mathbf{dQ_\Phi}$ matrix targeting individual modes. The gain and phase margins satisfied the limits, (\ref{eq79}) and (\ref{eq80}), as shown in Fig. \ref{fig19b}. Additionally, the disk-based margins were evaluated using MATLAB Robust Control Toolbox, and the minimum disk-based GM and PM were $\pm$9.33 dB and $\pm$52.4 deg, respectively. Variation of closed-loop eigenvalues with spool speed in Fig. \ref{fig20a} points to the closed-loop system being stable. The Bode plot of sensitivity, complementary sensitivity, and open-loop transfer functions in Fig. \ref{fig21} shows that the system also possessed high reference tracking capability and robustness in the typical engine control bandwidth of 10 rad/s \cite{bib64}. This was further validated by disturbance rejection testing. The peak sensitivity was 3.22 dB and was located outside of this bandwidth.

\subsection{Performance analysis} \label{sec6.4}

\subsubsection{Reference tracking} \label{sec6.4.1}
\begin{figure*}[!ht] 
    \centering
    \includegraphics[width=0.9\linewidth]{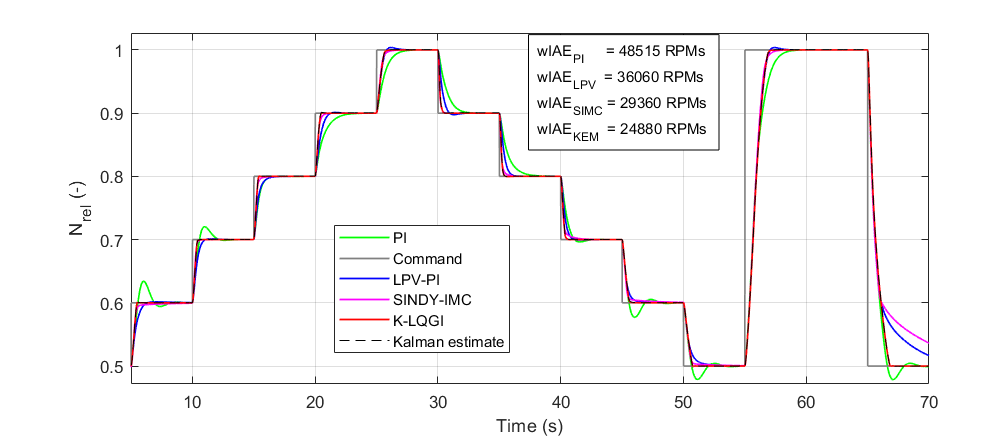}
    \caption{The control performance in sea-level conditions}
    \label{fig22}
\end{figure*}

\begin{figure*}[!ht] 
    \centering
    \begin{subfigure}[h]{0.4\textwidth}
        \centering
        \includegraphics[width=0.8\linewidth]{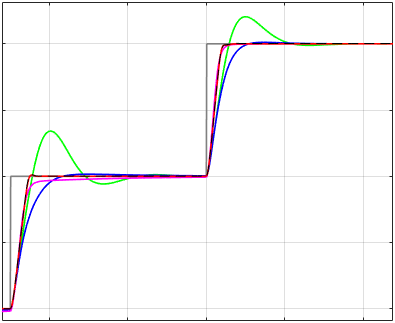}
        \caption{}
    \end{subfigure}
    \hspace{0mm}
    \begin{subfigure}[h]{0.4\textwidth}
        \centering
        \includegraphics[width=0.8\linewidth]{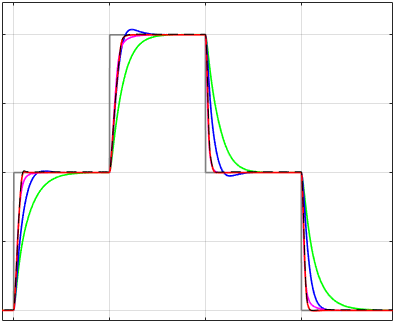}
        \caption{}
    \end{subfigure}
    \begin{subfigure}[h]{0.4\textwidth}
        \centering
        \includegraphics[width=0.8\linewidth]{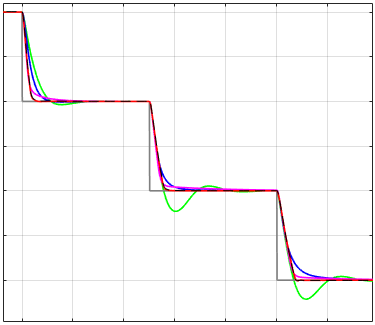}
        \caption{}
    \end{subfigure}
    \hspace{0mm}
    \begin{subfigure}[h]{0.4\textwidth}
        \centering
        \includegraphics[width=0.8\linewidth]{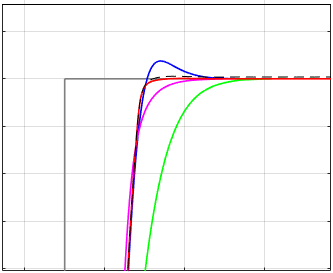  }
        \caption{}
    \end{subfigure}
    \caption{Details of certain response regions - (a) Small transient for low RPM acceleration, (b) high RPM, (c) low RPM deceleration, (d) large transient - acceleration. Solid lines were computed using the CLM without noise. The black dashed line is the estimation for noisy data using the Kalman observer}
    \label{fig23}
\end{figure*}

The performance of controllers was evaluated using the nonlinear CLM at sea level and varying flight conditions. The command profile was the same stair profile used for the optimization of K-LQGI in Fig. \ref{fig4}. The results for sea-level conditions are depicted in Fig. \ref{fig22} and details in Fig. \ref{fig23}. The wIAE in the figures stands for the weighted IAE in (\ref{eq71}). Since the LPV-PI and IMC controllers were based on only an approximation of the engine's nonlinear dynamics valid in the vicinity of the steady-state operating line, the different dynamical characteristics of acceleration and deceleration were handled poorly. By further increasing the LPV controller gains, the response was faster also during deceleration, but the overshoots became unacceptable during acceleration and small transients. Similarly, further decreasing the filter time constant $\tau_f$ of the IMC resulted in faster response at the cost of significantly increased sensitivity of the derivative term to the measurement noise.

Quantitatively, the K-LQGI controller with the Kalman observer exhibited the lowest wIAE. The performance during small transients is also of high significance, and the K-LQGI kept the same response characteristics, namely the speed and minimal overshoots, across the entire operating range, whereas the LPV-PI controller presented an overall slower response. The IMC controller got closest to the K-LQGI performance, but was also overall slower. Regarding the large transients, the initial phase was dominantly guided by the fuel flow limiters. Thus, the performance of the LPV-PI, IMC, and K-LQGI controllers was very similar, except for the deceleration to GI, where both LPV-PI and IMC controllers exhibited a sudden decrease in response speed, accompanied by a decrease in spool speed. This points to the nonlinear nature of the engine dynamics, further supported by the fuel flow limiting, which could not be captured by the scheduling approach. Since limiters were inactive during small transients, the fact that K-LQGI did not cause overshoots indicates the controller's ability to decrease the acceleration rate close to the desired state accordingly. The Kalman estimates showed the ability of KEM to predict the engine state reliably. 

\begin{figure}[!ht] 
    \centering
    \begin{subfigure}{0.4\textwidth}
        \centering
        \includegraphics[width=1\linewidth]{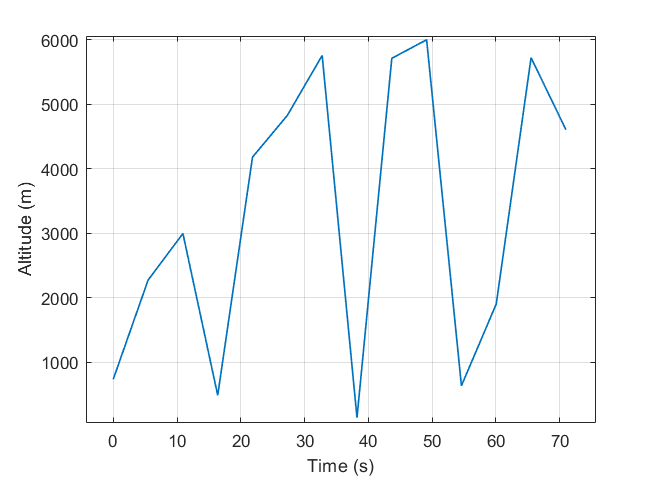}
        \caption{}
    \end{subfigure}
    \hspace{2mm}
    \begin{subfigure}{0.4\textwidth}
        \centering
        \includegraphics[width=1\linewidth]{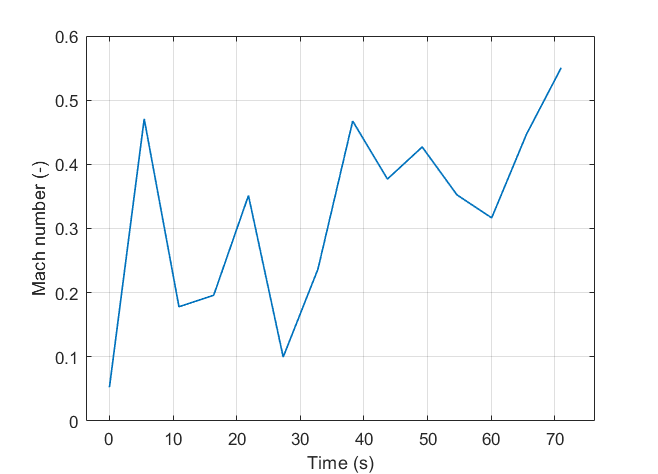}
        \caption{}
    \end{subfigure}
    \caption{(a) The varying flight altitude and (b) Mach number}
    \label{fig24}
\end{figure}

\begin{figure*}[!ht]
    \centering
    \includegraphics[width=0.9\linewidth]{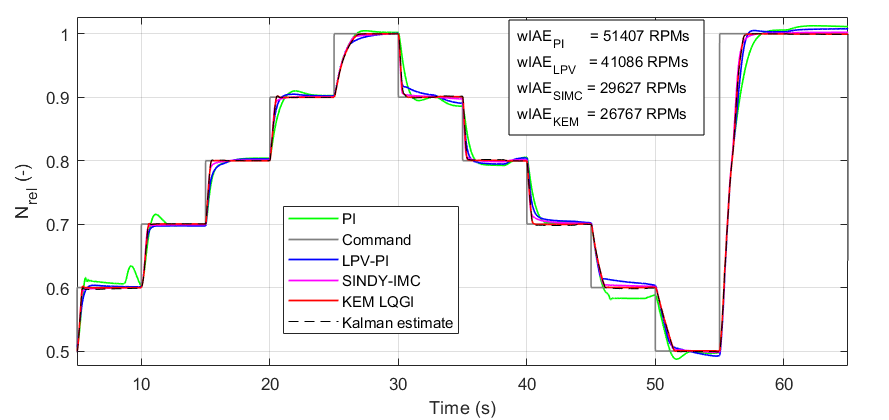}
    \caption{Response of the spool speed during varying flight conditions}
    \label{fig25}
\end{figure*}

Concerning the varying flight conditions, the altitude, $H$, and $M_0$ profiles are plotted in Fig. \ref{fig24}. These changes are unrealistic, but the main goal was to capture the effects of flight conditions during the 70-second simulation period. The results are depicted in Fig. \ref{fig25}. The classical and LPV PI controllers started to lose robustness, not being able to effectively reject the corresponding disturbances in engine behavior. The IMC performance was consistent but slightly slower compared to the K-LQGI. The deceleration to GI was not considered for the comparison during the higher $M_0$ of flight, as the minimum sustainable spool speed increases with $M_0$, and the simulation would leave the known region of the compressor map, leading to nonphysical simulation outputs. 

The wIAE, standard IAE, and maximum settling time (MST) are summarized in Table \ref{tab1}. The MST was defined as a time when the response last crossed the $N_d \pm 1 \%$ boundary during the full acceleration. The quantitative results also point to the superior performance of the K-LQGI controller, closely followed by the SINDy-IMC controller. 

\begin{table*}[!ht] 
\caption{Comparison of different performance metrics in the sea-level and varying flight conditions}\label{tab2}
\begin{tabular*}{\textwidth}{@{\extracolsep\fill}lcccccccc}
\toprule%
& \multicolumn{4}{@{}c@{}}{Sea-level conditions} & \multicolumn{4}{@{}c@{}}{Varying conditions} \\\cmidrule{2-5}\cmidrule{6-9}%
Controller & PI & LPV-PI & IMC & K-LQGI & PI & LPV-PI & IMC & K-LQGI \\
\midrule
wIAE (RPMs)  & 48515 & 36060 & 29360 & 24880 & 51407 & 41086 & 29627 & 26767 \\
IAE (RPMs) & 19115 & 18000 & 16882 & 13845 & 17508 & 15005 & 11722 & 11190 \\
MST\footnotemark[1] (s) & 3.17 & 1.95 & 2.13 & 1.9 & 3 & 2.75 & 2.21 & 1.89 \\
\end{tabular*}

\footnotetext[1]{Maximum settling time defined as a time when the response last crosses the $N_d\pm 1 \%$ boundary during acceleration from 0.5 to 1}
\label{tab1}

\end{table*}

\subsubsection{Disturbance rejection} \label{sec6.4.2}
\begin{figure}[!ht]
    \centering
    \begin{subfigure}{0.4\textwidth}
        \centering
        \includegraphics[width=1\linewidth]{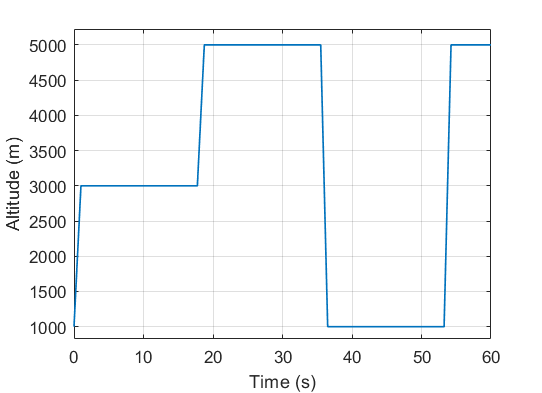}
        \caption{}
    \end{subfigure}
    \hspace{2mm}
    \begin{subfigure}{0.4\textwidth}
        \centering
        \includegraphics[width=1\linewidth]{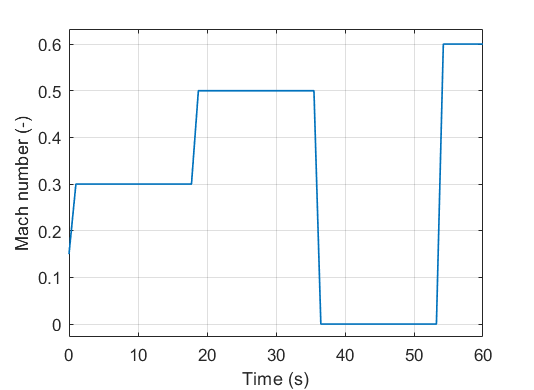}
        \caption{}
    \end{subfigure}
    \caption{Disturbances modeled by aggressive changes in (a) altitude and (b) Mach number}
    \label{fig26}
\end{figure}

\begin{figure}[!ht] 
    \centering
    \includegraphics[width=1\linewidth]{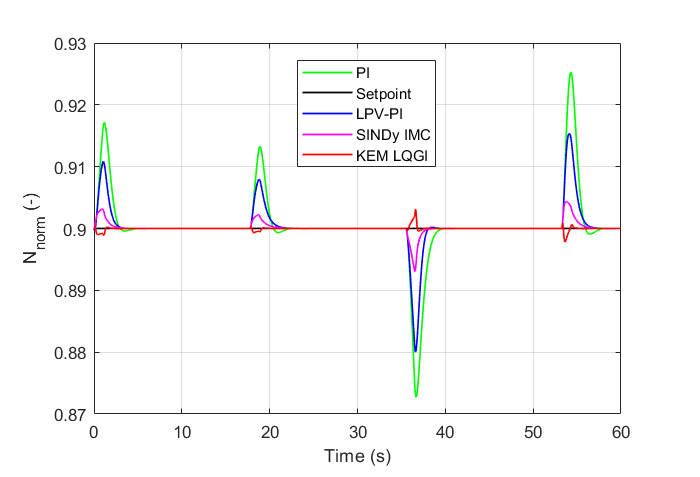}
    \caption{Disturbance rejection results. The K-LQGI controller exhibits the best disturbance rejection (red), followed by the SINDy-IMC (magenta)}
    \label{fig27}
\end{figure}

The most common disturbances acting on the engine are caused by the changing flight conditions - $H$ and $M_0$, and aircraft maneuvers affecting the air flow conditions, e.g., increased transient flow separation in the intake. Here, the disturbance rejection capability of the compared controllers was assessed for unrealistic abrupt changes in flight conditions, depicted in Fig. \ref{fig26}, leading to changes in the inlet total pressure and temperature. The setpoint was set to 90 $\%$ of the maximum spool speed as it often corresponds to the nominal operating point, where the engine spends most of the time. The results plotted in Fig. \ref{fig27} show that the K-LQGI controller exhibited the best disturbance rejection, with the SINDy-IMC close by. 

\section{Conclusion}\label{sec7}
To the author's best knowledge, based on a rather extensive literature review, this is the first paper utilizing the KOT for the design of an optimal GTE controller. Furthermore, it is also the first to combine the SINDy algorithm with the KOT for estimation of the GTE governing equation with a subsequent Koopman canonical transform to the KEM. Another contribution is the constrained optimization of a low-order set of eigenvalues using metaheuristic global optimization tools, e.g., GA and PSO, and the projection of the autonomous trajectories in the time domain, providing an optimal set of eigenvalues for the eigenfunction construction from data. The objective function rejects the unstable, too oscillatory, and too lightly damped eigenvalues. Moreover, the algorithm can switch from distinct eigenvalues to repeated ones when the distance of any two eigenvalues is smaller than a specified margin. The discovered set of eigenfunctions and eigenvalues can also be extended by exploiting the algebraic structure. However, by a product of two eigenfunctions, we obtain a new eigenfunction that always decays faster than the original two functions, and thus it contributes less to the dynamics.  

The KEM with the Kalman observer was capable of an accurate state prediction even outside the region of validity of the LPV system. Furthermore, it allows for more aggressive integral control balanced by targeting of individual modes, where the modes with the lowest damping are the most important ones. This is captured by the reduced overshoots compared to PI and LPV-PI controllers. Furthermore, the LPV-PI and IMC controllers both struggle with the large deceleration, which points to too small PI gains in the low-RPM region for the deceleration. However, further increases of these gains would cause larger overshoots during small accelerations for the LPV-PI controller and increased sensitivity to noise for the IMC controller. Thus, the locally linearized controllers are overall effectively outperformed by the optimized KEM LQGI. 

These methods can be extended to an adaptive framework. For the KEM, the eigenfunctions can be fixed, and the matrix of Koopman modes, $\mathbf{C}$, can be adapted in real time using the ridge regression for a finite sequence of past spool speed measurements. Also, the control law can be further replaced, e.g., by a nonlinear MPC controller. Both the identification method and the subsequently designed controller proved to be efficient for the control of a single-spool turbojet GTE. For multi-spool turbojet and turbofan GTEs, the proposed eigenvalue optimization needs to be modified. This is currently the subject of our ongoing research. The issue is primarily the ambiguity in coefficients $\textbf{K}$ encoding both the unknown initial conditions of eigenfunctions and the mode amplitudes. These cannot be easily decoupled like in the case of 1st-order systems, where the variation of $\textbf{K}$ for different trajectories is caused solely by the changing initial conditions of $\mathbf{\Phi}$. Fortunately, the presented eigenfunction identification procedure works in higher dimensions as well. 

In future research, we will extend this framework to multi-rotor and turbofan engines and test other control strategies such as the MPC. 

\backmatter

\bmhead{Acknowledgements}

\bmhead {Author contribution}
The author confirms sole responsibility for the following: study conception and design, simulation and data collection, analysis and interpretation of results, and manuscript preparation.

\bmhead {Funding}
The author declares that this research was supported by funding provided by the government of the Czech Republic.

\bmhead{Data availability}
Data will be made available on reasonable request.

\section*{Declarations} \label{sec8}

\bmhead {Conflict of interest}
The author has no conflicts of interest and no relevant financial or non-financial interests to disclose.

\begin{appendices}

\section{Parameter-affine LPV approximation of the input dynamics}\label{secA1}
The input dynamics given by $\nabla\mathbf{\Phi}(N) g(N)$ are relatively complex as each of the entries is given by a product of two linear combinations of nonlinear basis functions, which, in the case of $\nabla\mathbf{\Phi}$, are even different for each entry. Thus, it would be suitable to approximate this complicated expression by an affine LPV representation, accelerating the computation of the Koopman observer equations for real-time applications.

The goal is to decompose the input dynamics as
\begin{equation} \label{eqA1}
    \nabla\mathbf{\Phi} (N)g(N) = \sum_{k=1}^m \rho_k(N) \mathbf{\tilde{G}_k}=\mathbf{\tilde{G}}\boldsymbol{\rho},
\end{equation}
where $\mathbf{\tilde{G}} = \begin{bmatrix} \mathbf{\tilde{G}_1} & \mathbf{\tilde{G}_2} & \cdots & \mathbf{\tilde{G}_m} \end{bmatrix}$ are basis matrices of the decomposition and $\boldsymbol{\rho} = \begin{bmatrix} \rho_1 & \rho_2 & \cdots & \rho_m\end{bmatrix}^\intercal$ are nonlinear scalar scheduling functions. 

Let us denote $\textbf{G}=\nabla\mathbf{\Phi} (N)g(N)$. The primary goal is to optimize $\boldsymbol{\rho}$, parameterized by $\textbf{p}$, solving the following minimization problem: 
\begin{align} \label{eqA2}
    \textbf{p} &= \arg \min_{\textbf{p}} \| \textbf{G} - \mathbf{\tilde{G}}\boldsymbol{\rho}(N,\textbf{p}) \|_F^2 \ , \nonumber \\
    \mathbf{\tilde{G}} &= \big(\boldsymbol{\rho}(N)\boldsymbol{\rho}(N)^\intercal -\alpha_g \textbf{I}\big)^{-1}\mathbf{\boldsymbol{\rho}}(N) \mathbf{G^T} \hspace{2mm} \bigg |_{\mathbf{p}=\text{const.}}\ , 
\end{align}
where $\textbf{p}$ is the vector of nonlinear parameters of the scheduling functions and $\alpha_g$ is the regularization parameter.

For the discrete-time system, the matrices can be rewritten as
\begin{align}
    \mathbf{L} &= e^{\mathbf{\Lambda} \Delta t}, \\ \mathbf{G_d} &= \mathbf{\Lambda^{-1}(L-I)\tilde{G}\boldsymbol{\rho}} \nonumber \\
    &= \mathbf{\tilde{G}_d\boldsymbol{\rho}},
\end{align}
where $\mathbf{\tilde{G}_d=\Lambda^{-1}(L-I)\tilde{G}}$ is a new basis matrix and vector $\boldsymbol{\rho}$ is evaluated in discrete time steps $t_k=k\Delta t$.

This problem can be handled with known optimization solvers, e.g., GD, GA, or PSO. The latter two are capable of a quick search in the parametric space, whereas the GD can be utilized to analytically fine-tune the solution. However, the linear coefficients would need to be optimized as well, leading to high-dimensional gradients and slower overall performance. Thus, the meta-heuristic global optimization tools might be preferred. 

\begin{figure}[!ht]
    \centering
    \includegraphics[width=1.05\linewidth]{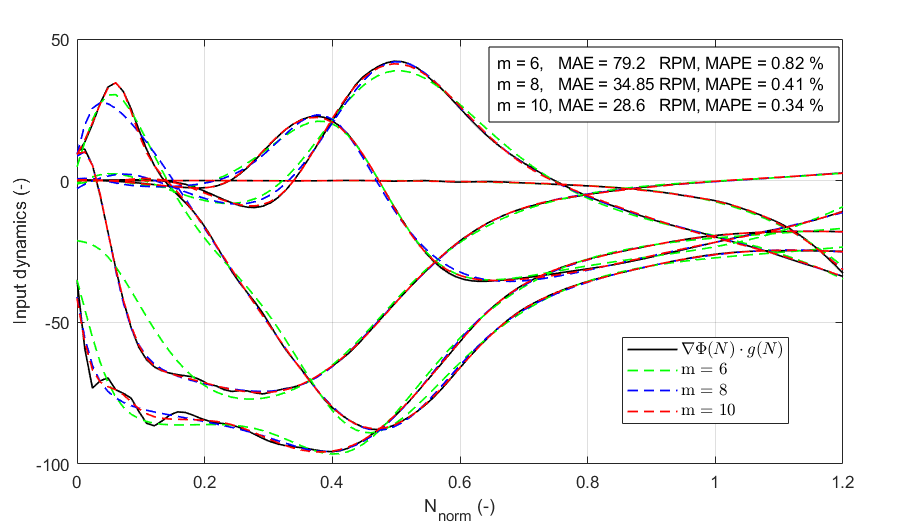}
    \caption{Comparison of affine LPV fitting for different numbers of LFs as basis functions}
    \label{figA1}
\end{figure}

The results of LPV decomposition with KEM prediction MAE, using different numbers of LFs, are depicted in Fig. \ref{figA1}. It can be concluded that the approximation is sufficient for $m = 8$. 

The computation of the input dynamics in each time step reduces only to an evaluation of 8 nonlinear functions and further matrix operations, which are rather fast. This is an improvement compared to the evaluation of $6 \times 10$ functions for the gradient and 5 functions for $g(N)$. The model can also be converted to the polytopic form. 

\end{appendices}

\bibliography{sn-bibliography}

\end{document}